\documentclass[letterpaper]{article} 
\usepackage{aaai2027}  
\makeatletter
\renewcommand{\copyright@year}{}
\renewcommand{\copyright@text}{}
\makeatother
\usepackage[hyphens]{url}  
\usepackage{graphicx} 
\usepackage{natbib}  
\usepackage{caption} 
\usepackage{algorithm}
\usepackage{algorithmic}

\usepackage{newfloat}
\usepackage{listings}
\DeclareCaptionStyle{ruled}{labelfont=normalfont,labelsep=colon,strut=off} 
\floatstyle{ruled}
\newfloat{listing}{tb}{lst}{}
\floatname{listing}{Listing}

\usepackage{booktabs}

\usepackage{multirow}
\usepackage{makecell}
\usepackage{adjustbox}

\usepackage{verbatim}
\usepackage[T1]{fontenc}
\usepackage{amsfonts}  
\usepackage[table]{xcolor}
 \usepackage{tikz}
\usetikzlibrary{shapes, arrows, positioning, fit}
 
\usepackage{siunitx}  

\usepackage{array}
\usepackage{tabularray}
\usepackage{amsmath}
\usepackage{amssymb}
\usepackage{threeparttable}

\usepackage{tabularx}

\definecolor{highlight}{gray}{0.93}
\definecolor{securegateblue}{rgb}{0.9, 0.95, 1.0}

\usepackage{subcaption}

\usepackage{colortbl}

\definecolor{lime}{rgb}{0.88,2,10}

\usepackage[spaces,hyphens]{xurl}
\newcommand*{\Resize}[2]{\resizebox{#1}{!}{$#2$}}%
\usepackage{textcomp}

\newcommand{\fref}[1]{Fig.~\ref{#1}}

\title{Resource-Aware Federated Mixture-of-Experts with Adaptive Pruning for Onboard 
{Learning in LEO Satellite Constellations}}
\author{
Mohamed Shaaban, Mohamed Elmahallawy, Marius Bernahrndt, Tobias Hecking
}
\affiliations{School of Engineering and Applied Science, Washington State University, Richland, WA 99354, USA\\Institute for Software Technology, German Aerospace Center (DLR), Porz, 51147 Cologne, Germany
\\Email:\{mohamed.shaaban, mohamed.elmahallawy\}@wsu.edu, \{marius.bernahrndt, tobias.hecking\}@dlr.de}

\begin{document}

\maketitle

\begin{abstract}

Low-Earth-orbit (LEO) satellites are increasingly expected to perform onboard learning for applications such as disaster response and environmental monitoring. However, conventional federated learning (FL) is ill-suited to onboard satellite learning, as it assumes computational, memory, and communication resources beyond the capabilities of resource-constrained LEO platforms, often necessitating the transmission of raw imagery to ground stations. We present \textsc{COSMIC-FL}, a resource-aware FL framework for efficient onboard learning in LEO satellite constellations. \textsc{COSMIC-FL} introduces two complementary Mixture-of-Experts (MoE) architectures: a \emph{Sliced} design that shares backbone representations while activating task-specific channel subsets, and a \emph{Modular} design that employs lightweight gating to route inputs to physically separated expert networks. A semantic class-to-expert mapping enables each satellite to train, update, and communicate only the expert paths relevant to its local data. To further improve efficiency, \textsc{COSMIC-FL} integrates staged optimization with three structured pruning strategies: \emph{server-side pruning}, \emph{client-side fixed-ratio pruning} with mean-vote aggregation, and \emph{adaptive client-side per-layer pruning} based on aggregated importance and a MAD-based gap criterion. Combined with semantic expert routing, these techniques jointly adapt computation and model sparsity to both data semantics and layer importance, yielding a favourable accuracy--efficiency trade-off for heterogeneous space platforms. Extensive experiments on six image classification benchmarks under highly non-i.i.d.\ settings demonstrate that \textsc{COSMIC-FL} maintains competitive accuracy while reducing communication, computation, and energy consumption by up to 80\% compared with SOTA FL methods. We further validate \textsc{COSMIC-FL} on an NVIDIA Jetson AGX Orin, where measurements of latency, throughput, GPU memory usage, power, and temperature confirm that the analytical efficiency gains translate to realistic embedded deployments.

\end{abstract}

\section{Introduction}\label{sec:intro}

Recent years have witnessed rapid growth in low-Earth-orbit (LEO) satellite constellations (e.g., nanosatellites) for Earth observation and communication. These platforms capture massive volumes of {\em high-resolution imagery}~\cite{wang2023satellite} to be processed using machine learning (ML). However, centralized learning is impractical: contact windows with ground stations (GSs) are brief, and downlink bandwidth is limited. In addition, satellite imagery often carries commercial or privacy restrictions that prevent unrestricted transmission~\cite{elmahallawy2023secure}. Federated learning (FL)~\cite{mcmahan2017communication} offers a promising alternative, allowing each satellite to train locally on-board while sharing only model updates. In principle, FL enables satellites to collaboratively learn deep models without transmitting raw data to Earth, while preserving data privacy.

Despite this potential, deploying FL for {\em onboard} learning on small satellites remains challenging. CubeSats operate under strict compute, memory, and power constraints, and flight-grade processors typically provide only a few GFLOPS within tight energy budgets. Network connectivity is intermittent and asymmetric, and contact windows with ground stations are short and unpredictable~\cite{elmahallawy2024communication}. These constraints make conventional deep-learning workloads difficult to execute in orbit. Moreover, standard FL algorithms such as FedAvg~\cite{li2019convergence} assume homogeneous devices, ample compute, and synchronous updates—assumptions that rarely hold in LEO constellations. Recent satellite-FL frameworks—FedSN~\cite{lin2024fedsn}, FedLEO~\cite{elmahallawy2023optimizing}, and FedSat-LAM~\cite{pasandi2025fedsat}—improve convergence under intermittent LEO links through pseudo-/hierarchical aggregation, intra-plane propagation, sink scheduling, and multi-hop offloading, but they still mostly rely on  monolithic models rather than sparse or expert-based architectures. 

In response, we propose \mbox{\textsc{COSMIC-FL}} ({\bf Co}llaborative {\bf S}atellite {\bf MI}xture-of-Experts {\bf C}ompression-{\bf FL}), a framework for {\it onboard} training of ML models in satellite constellations by combining sparse {Mixture-of-Experts (MoE)} architectures with {\it multi-level} model pruning. \mbox{\textsc{COSMIC-FL}} offers two MoE designs: {\bf \em Sliced-MoE}, which partitions the model into channel slices across satellites, and {\bf \em Modular-MoE}, where experts specialize in subtasks or regions. 
A semantic class-to-expert mapping assigns each training sample to its single responsible expert, so each satellite trains only one expert at a time; a lightweight gate, trained to mirror this mapping, weighs the expert outputs at inference.

To further improve efficiency, we introduce three pruning strategies to remove entire filters or channels, producing compact subnetworks suitable for resource-constrained LEO hardware. (i) {\em centralized weight-based pruning} at the ground station using global importance scores; (ii) {\em client-informed fixed-ratio pruning} based on aggregated layer-wise importance summaries; and (iii) {\em client-informed adaptive pruning} that adjusts pruning strength per layer. Overall, our framework reduces computation, memory, and communication overhead while preserving the capacity of a larger expert model. Our main contributions are as follows:

\begin{itemize}

\item 
{\textbf{Resource-Aware Onboard Training:} We propose \textsc{COSMIC-FL}, an efficient FL framework for resource-constrained LEO satellites that optimizes and communicates only phase-relevant parameters through staged expert specialization, gate fine-tuning, and head fine-tuning.}

\item 
{\textbf{Complementary Expert Implementations:} We design two MoE variants with different resource trade-offs: Sliced-MoE with shared tensors and channel-based expert paths, and Modular-MoE with physically separated expert towers, enabling analysis of parameter efficiency versus runtime performance.}

\item 
\textbf{Federated Structured Pruning:} We develop three structured pruning strategies: {\em centralized weight-based, satellite-informed fixed-ratio, and adaptive pruning}, where the latter uses a MAD-based gap criterion for data-driven, gap-constrained compression.
\item 
{\textbf{Server-Side and Embedded Evaluation:}
We evaluate \textsc{COSMIC-FL} and the baselines on six datasets under i.i.d.\ and non-i.i.d.\ partitions. We further execute the proposed training pipeline on an NVIDIA Jetson AGX Orin as an embedded feasibility study and provide a separate inference characterization for FedAvg, FedProx, PM-MoE, and the proposed variants on the same device.}

\end{itemize}

\section{Related Work}\label{sec:rlt_wrk}

\noindent{\bf Federated Learning for Satellite Networks.}
Recent works adapt FL to satellite environments with challenges such as intermittent connectivity and limited bandwidth. Matthiesen {\em et al.}~\cite{matthiesen2023federated} survey satellite FL challenges, while FedLEO~\cite{elmahallawy2023optimizing}, DEDFL~\cite{zhang2025ground}, and satellite adaptations of FedAvg/FedProx~\cite{kim2026bringing} improve convergence through propagation, scheduling, and communication optimization. However, these approaches mainly address connectivity and orchestration, assuming models remain feasible for resource-constrained satellites.

\noindent{\bf Mixture-of-Experts in Federated Learning.}
MoE-based FL methods improve personalization under non-i.i.d.\ data. PM-MoE~\cite{feng2025pm}, pFedMoE~\cite{yi2026pfedmoe}, dFLMoE~\cite{xie2025dflmoe}, and ShiftEx~\cite{bhope2025shift} leverage expert specialization and routing to handle heterogeneity. However, they primarily optimize accuracy and personalization, increasing model complexity and limiting applicability to resource-constrained LEO platforms.

\noindent{\bf Model Pruning in Federated/Edge AI.}
Pruning has been explored to reduce FL communication and computation costs. Li {\em et al.}~\cite{li2024model}, FedSpaLLM~\cite{bai2025fedspallm}, AutoFLIP~\cite{interno2024adaptive}, and MP-FSL~\cite{jia2024model} investigate sparse aggregation, adaptive pruning, and channel reduction. However, these methods target terrestrial edge settings and do not jointly optimize pruning with adaptive expert selection for satellite FL.

{\em Unlike existing approaches, COSMIC-FL jointly enables sparse expert collaboration and structured compression for resource-constrained satellite constellations through federated MoE training and adaptive pruning.}


\begin{figure*}[!t]
\centering
\includegraphics[width=\linewidth]{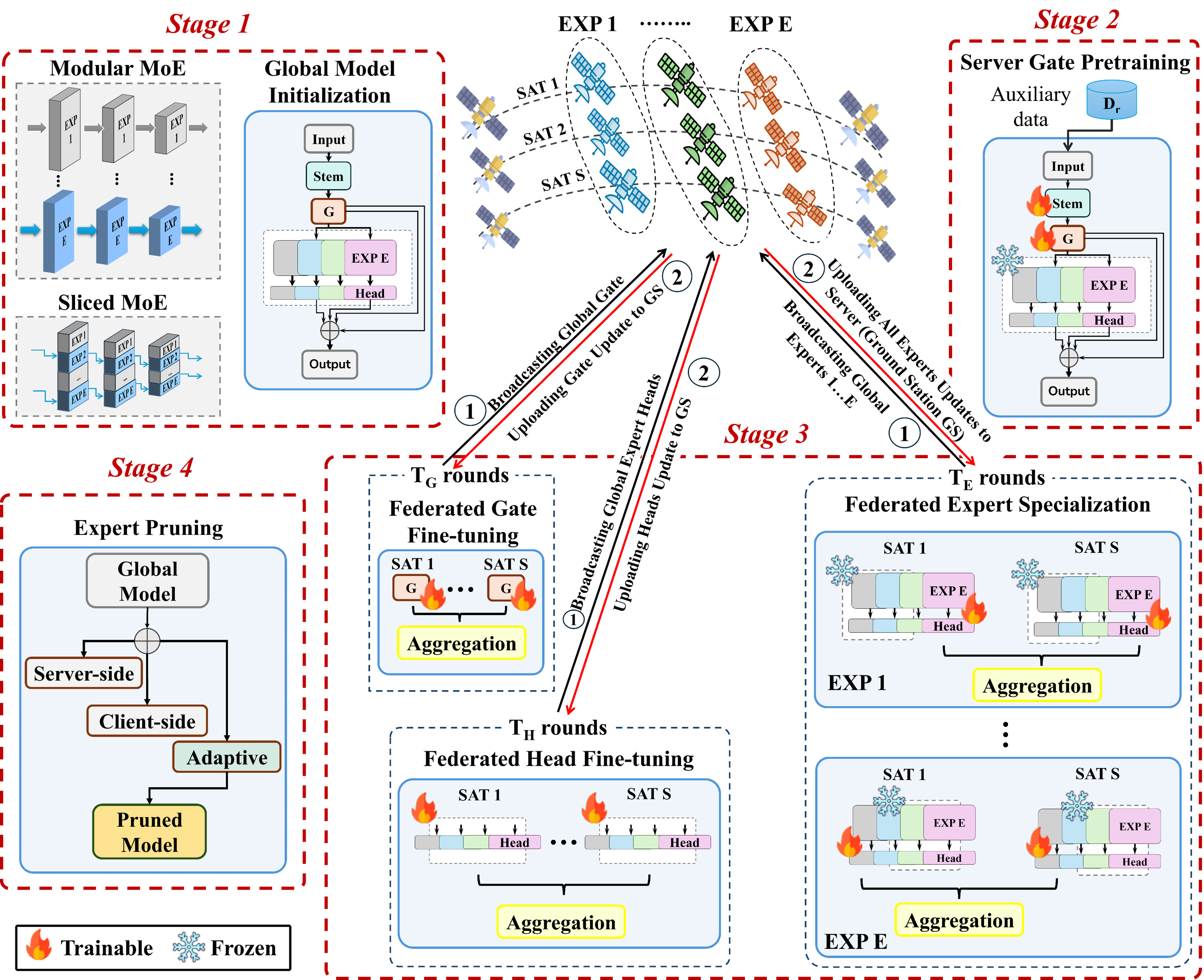}
\caption{Overview of the four stages of the \textsc{COSMIC-FL} framework.}
\label{fig:cosmicfl_framework}
\end{figure*}

\section{Methodology}\label{sec:method}

This section presents \textsc{COSMIC-FL}, a framework for practical onboard ML training in LEO satellite constellations. \textsc{COSMIC-FL} consists of four key components. First, it introduces two complementary MoE architectures: \emph{Sliced-MoE}, which shares convolutional representations across channel-based expert paths for parameter efficiency, and \emph{Modular-MoE}, which uses physically separated expert towers for independent specialization and efficient dense execution with structured pruning.
Second, a lightweight gating module performs semantic class-to-expert routing, enabling each satellite to update only the expert associated with its local samples while combining expert outputs through softmax weighting during inference. Third, the federated training protocol is divided into communication-efficient stages to accommodate intermittent connectivity and limited bandwidth. Finally, we introduce three pruning strategies—\emph{server-side global pruning}, \emph{fixed-ratio client-side pruning}, and \emph{adaptive client-side pruning}—that further reduce model size by adapting per-layer sparsity using aggregated client importance scores.  The overall \textsc{COSMIC-FL} framework is illustrated in \fref{fig:cosmicfl_framework}.
\subsection{Mixture-of-Experts Architectures} \label{sec:moe-arch}

We propose two complementary MoE architectures for FL in LEO satellite networks: \emph{Sliced-MoE} for parameter-efficient collaboration and \emph{Modular-MoE} for independent expert specialization. Both use a lightweight gating mechanism for distributed expert routing. Each architecture consists of four core components:
\begin{enumerate}
    \item \textbf{Shared Backbone (Stem)}: Initial convolutional layers that process raw inputs into a unified feature map $z = h(x)$, providing consistent low-level representations accessible to both the gating module and all experts.
    
\item \textbf{Gating Module}: A lightweight network $g(\cdot)$ that maps features $z$ to expert logits. Routing weights are given by softmax as $\label{eq:gate-softmax}
    p_e(x) = \frac{\exp(z_e)}{\sum_{j=1}^{E} \exp(z_j)}, \quad e \in \{1,\ldots,E\}$, with $E$ is the number of experts, $z_e$ is the $e$-th component of $g(z)$ (the gate logit for expert $e$), and $p_e(x)$ is the routing weight of expert $e$.

\item \textbf{Experts:} Specialized sub-networks (paths) that focus on particular subsets of classes or data patterns. During training, each expert processes only the samples routed to it, allowing it to learn task-specific representations. This design promotes expert specialization while improving computational efficiency by reducing the number of parameters and computations required per path.
    
    \item \textbf{Expert-Specific Heads}: Dedicated linear classifiers per expert, mapping expert-processed features to class-specific logits.
\end{enumerate}

\subsubsection{Sliced Mixture-of-Experts.} We propose a \emph{Sliced-MoE} architecture that enables parameter-efficient collaboration in resource-constrained satellite FL. Unlike conventional MoE designs that allocate separate parameters to each expert, Sliced-MoE partitions intermediate feature channels of a shared convolutional block into disjoint slices assigned to different experts. This allows experts to specialize while reusing the same underlying model weights, significantly reducing memory and communication overhead.

Let the shared convolutional block produce a feature map $z \in \mathbb{R}^{C \times H \times W}$ with $C$ channels. The channels are partitioned into $E$ disjoint index sets $\mathcal{I}_1, \ldots, \mathcal{I}_E$ such that $\mathcal{I} = \bigcup_{e=1}^{E}  \{1,\ldots,C\}$ and $\mathcal{I}_i \cap \mathcal{I}_j = \emptyset$ for $i \neq j$. Expert $e$ receives only the slice corresponding to $\mathcal{I}_e\subseteq \mathcal{I}$ of channels: $z^{(e)} = z[\mathcal{I}_e] \in \mathbb{R}^{|\mathcal{I}_e| \times H \times W}$, where $|\mathcal{I}_e|$ is the number of channels assigned to expert $i$. Equivalently, this slicing can be implemented via element-wise masking:
\begin{equation}
z^{(e)} = \mathbf{m}_e \odot z, \quad \mathbf{m}_e \in \{0,1\}^{C \times 1 \times 1},
\end{equation}
where $\mathbf{m}_e$ is the channel mask for expert $e$ (ones at positions in $\mathcal{I}_i$, zeros elsewhere) and $\odot$ denotes the Hadamard (element-wise) product.

The binary-mask notation is a conceptual representation of channel membership. In the implementation, expert paths are represented through explicit channel ranges or index sets. After structured pruning, channels that are no longer used by any expert path are physically removed from the shared convolutional layers, and the remaining path indices are remapped to the compacted channel coordinates.

By sharing the same convolutional weights while activating only expert-specific channel subsets, \textit{Sliced-MoE} enables satellites to update different experts without replicating the full model.  During federated rounds, satellites assigned to expert $i$ (through masking) compute gradients only for channels in $\mathcal{I}_e$, enabling lightweight expert updates,  reducing computation and communication while preserving the overall capacity of the global model.

\subsubsection{Modular Mixture-of-Experts.}\label{sec:MoE_modular}

Unlike \emph{Sliced-MoE}, which shares convolutional tensors across expert paths, \emph{Modular-MoE} assigns each expert a physically separated tower with dedicated parameters, layers, and classification heads. While this increases the static parameter footprint, it enables simpler dense execution and independent expert training, communication, and structured pruning.


Without loss of generality, \textit{COSMIC-FL} primarily adopts the \emph{Modular-MoE} architecture because it (i) enables expert-wise updates and pruning without affecting other experts, and (ii) remains compatible with standard dense training and inference operators. In this design, each expert tower $f_i(\cdot)$ is implemented as an independent ML model composed of an input convolution layer followed by several residual blocks (see \fref{fig:cosmicfl_framework}). The residual connections stabilize optimization by learning residual corrections around identity mappings~\cite{he2016deep}. Each expert also includes its own classification head $c_i(\cdot)$.
The final prediction is computed as {a dense, gate-weighted combination of all $E$ experts such that $\hat{y} = \sum_{i=1}^{E} p_i(x)\; c_i\!\left(f_i(z)\right),$ where $z = h(x)$ is the feature representation produced by the shared backbone, and $p_i(x)$ is the routing weight assigned to expert $i$ as defined in~\eqref{eq:gate-softmax}. {For \emph{Sliced-MoE}, this dense combination does not increase inference cost with $E$, since the experts partition a fixed total channel width.}

\subsection{Semantic Class-to-Expert Mapping and Gate Initialization}\label{sec:moe-class-mapping}

To enable efficient expert specialization, \textit{COSMIC-FL} assigns classes to experts in a semantically coherent manner. The goal is to cluster classes such that those handled by the same expert are semantically similar (high within-expert similarity), while classes assigned to different experts remain well separated (low between-expert similarity). To achieve this, we compute a class-to-expert mapping using an auxiliary dataset $D_r$, following the idea of leveraging a reserved dataset to improve initial routing~\cite{jiang2025heterogeneous}. 

We first employ a pretrained CNN feature extractor $\phi$ to map inputs into a semantic feature space. For each class $m \in \{1,\ldots,M\}$, we compute a class centroid using samples from the auxiliary dataset as
$\boldsymbol{\mu}_m = \frac{1}{|D_r^m|} \sum_{x \in D_r^m} \phi(x)$, where $D_r^m \subseteq D_r$ contains samples with label $m$. The set of centroids $\{\boldsymbol{\mu}_1, \ldots, \boldsymbol{\mu}_M\}$ is then partitioned into $E$ clusters using $K$-means:
\begin{equation}
\displaystyle
\Resize{7.6cm}{\min\limits_{\{\mathcal{G}_1,\ldots,\mathcal{G}_E\}}\sum\limits_{e=1}^{E} \sum\limits_{m \in \mathcal{G}_e} \big\| \boldsymbol{\mu}_m - \mathbf{v}_e \big\|^2,
~
\mathbf{v}_e = \frac{1}{|\mathcal{G}_e|} \sum\limits_{m \in \mathcal{G}_e} \boldsymbol{\mu}_m.}
\end{equation}
Each group $\mathcal{G}_e$ is assigned to one expert, yielding the mapping $\pi(m) = e$ if $m \in \mathcal{G}_e$. This mapping guides sample routing during expert specialization and provides a semantically meaningful initialization for the gating mechanism.

To stabilize routing before federated training, we pre-train the gating module using the auxiliary dataset $D_r$. During this phase, only the gate and shared backbone are optimized, while all expert parameters remain frozen. This server-side pre-training establishes informative feature representations and reliable initial routing decisions, facilitating subsequent federated expert specialization.


\subsection{Federated Learning Training Stages} \label{sec:moe-training}


\subsubsection{Federated Expert Specialization via Rotation.}

After gate initialization (Section~\ref{sec:moe-class-mapping}), expert specialization is performed through a federated \emph{expert-rotation} phase over $T_E$ communication rounds. The schedule is independent of the number of experts $E$, since all experts are updated within each round. At each round, the GS broadcasts the current expert and classification-head parameters to available satellites, while the shared backbone and gate remain fixed after initialization. In \emph{Sliced-MoE}, each expert corresponds to a channel slice of shared convolutional tensors, whereas in \emph{Modular-MoE}, each expert is an independent tower. Each satellite updates only the active expert and its classification head in a randomized order, while other components remain frozen. The frozen backbone and gate participate only in forward propagation and incur no gradient, optimizer-state, or communication cost during this phase.

For expert $e$, each satellite trains on local samples assigned by the class-to-expert mapping $\pi$, i.e., samples satisfying $\pi(s)=e$, and uploads only the updated expert parameters. The GS aggregates each expert independently using federated averaging: $
\theta_e^{(t+1)}=\frac{1}{|\mathcal{S}|}\sum_{s\in\mathcal{S}}\theta_{s,e}^{(t+1)},$
where $\theta_{s,e}^{(t+1)}$ denotes the local update of expert $e$ from satellite $s$. Since all experts are updated in every round, the total rotation duration remains fixed at $T_E$ regardless of $E$; increasing the number of experts only affects per-round computation and communication.


Training experts independently in this rotating manner {\em significantly reduces the computation and communication burden on each satellite}, since only a small portion of the model is  updated and exchanged at a time: gradients, optimizer state, and uploads are confined to one expert and its head per phase, and with sparse expert training a satellite trains and uploads only the experts covering its local classes. The result of this phase is a set of specialized experts that capture task- or class-specific knowledge, which are later refined during the gate and head fine-tuning stages.


\subsubsection{Federated Gate Fine-Tuning}
 

After expert specialization, the gating module is fine-tuned so that routing decisions align with the trained experts. In each communication round $t_G$, the GS broadcasts the current global gate to the available satellites. Each participating satellite updates only the gate parameters using its local data, while keeping the experts, their classification heads, and the shared backbone frozen. The updated gate parameters are then sent back to the GS, which aggregates them using federated averaging. This process is repeated until routing becomes consistent with the learned expert specializations.

The gate maps the shared feature representation $z$ to expert logits $z_e$, which are converted into routing weights via the softmax function~\eqref{eq:gate-softmax} as $p_e(x) = \frac{\exp(z_e)}{\sum_{j=1}^{E} \exp(z_j)}, \quad e \in \{1,\ldots,E\}$, with $z_e$ defines the gate logit corresponding to expert $e$. These routing weights combine the outputs $\mathbf{o}_e$ of expert $e$ into the final routed output: 
$\mathbf{o} = \sum_{e=1}^{E} p_e(x)\,\mathbf{o}_e$.

During this phase, satellites optimize only the gate parameters while all expert networks and classification heads remain fixed. The gate is trained using local data to learn routing decisions consistent with the class-to-expert mapping. After local training, the GS aggregates the updated gate parameters across participating satellites as $
\theta_{\mathrm{gate}}^{(t+1)} = \frac{1}{|\mathcal{S}|} \sum_{c \in \mathcal{S}} \theta_{s,\mathrm{gate}}^{(t+1)}$, with $\theta_{s,\mathrm{gate}}^{(t+1)}$ denotes the gate parameters updated by satellite $s$ and $\mathcal{S}$ is the set of participating satellites.



\subsubsection{Federated Head Fine-Tuning.}
In the final stage, the feature extraction and routing components are frozen, including the gating module, shared backbone, and expert convolutional blocks. Only the expert classification heads are optimized. At each round, the GS broadcasts the current heads to participating satellites, which perform local updates on their head parameters while keeping all other components fixed. The GS then aggregates the updated heads using federated averaging:
$
\theta_{\mathrm{head}}^{(t+1)}=
\frac{1}{|\mathcal{S}|}\sum_{s\in\mathcal{S}}\theta_{s,\mathrm{head}}^{(t+1)},
$
where $\theta_{s,\mathrm{head}}^{(t+1)}$ denotes the locally updated heads at satellite $s$. Local optimization minimizes cross-entropy loss with label smoothing using AdamW with gradient clipping and weight decay. By refining only the classification heads after expert specialization and routing stabilization, this stage consolidates knowledge across experts and produces the final global model.
\noindent\textbf{Scalability in experts and classes.} \textsc{COSMIC-FL} decouples the number of experts ($E$) from the number of classes ($C$). The communication rounds remain fixed at $T_E+T_G+T_H$ regardless of $E$, while the per-round communication grows only modestly: \emph{Sliced-MoE} maintains a nearly constant convolutional payload by partitioning shared channels, whereas \emph{Modular-MoE} scales linearly with $E$. The gating module incurs only $\mathcal{O}(E)$ overhead, and only the expert classification heads scale with $C$. Furthermore, sparse expert training allows each satellite to optimize and communicate only experts covering its local classes, reducing communication by up to 89\% with 100 clients (Supplementary Material~F). Empirically, $E{=}3$ consistently provides the best accuracy--efficiency trade-off across datasets with 10--100 classes (Supplementary Material~F).

\subsection{Structured (Physical) Expert Pruning}\label{subsubsec:pruning}

This section describes how \textsc{COSMIC-FL} compresses expert paths after dense training. Rather than introducing sparse masks, our goal is to obtain physically smaller subnetworks with reduced parameters, FLOPs, communication payloads, and memory footprint. We therefore adopt structured filter/channel pruning, which removes complete channels and reconstructs convolutional and batch-normalization layers with smaller dense tensors. Unlike unstructured pruning, structured pruning preserves compatibility with standard dense operators without requiring specialized sparse kernels.

For a convolution with kernel size $K\times K$ and output feature map size $H_{\text{out}}\times W_{\text{out}}$, the computational cost scales approximately as
\begin{equation}
\mathcal{O}\left(H_{\text{out}}W_{\text{out}}K^2C_{\text{in}}C_{\text{out}}\right),
\end{equation}
while the parameter count scales as $
\mathcal{O}\left(K^2C_{\text{in}}C_{\text{out}}\right).$
Thus, pruning output channels reduces the current layer width $C_{\text{out}}$ and, after physical layer reconstruction, also reduces the input width $C_{\text{in}}$ of subsequent layers. This cascading effect explains why structured channel pruning can reduce both model size and dense FLOPs, especially when applied consistently across consecutive layers.

For each convolutional layer, the importance of an output filter $W_i \in \mathbb{R}^{C_{\text{in}}\times K\times K}$ is quantified using its $L_1$ magnitude:
\begin{equation}
S_i = \lVert W_i\rVert_1 =
\sum_{c=1}^{C_{\text{in}}} \sum_{h=1}^{K} \sum_{w=1}^{K}
\left|W_{i,c,h,w}\right|.
\end{equation}
Given a set of retained filters $\mathcal{I}_{\text{keep}}$, we apply physical pruning by re-instantiating smaller convolutional and batch-normalization layers and copying the retained parameters by index selection. Since the expert towers contain residual blocks, pruning decisions must preserve tensor-shape compatibility at residual additions~\cite{he2016deep}. To avoid destabilizing early feature learning, pruning is applied only after an initial dense-training phase, following the common train--prune--fine-tune paradigm~\cite{han2015deep}.

\subsubsection{Federated Pruning Strategies.}\label{subsubsec:fed-pruning}
We introduce three federated pruning strategies that differ in the signals used to derive pruning decisions. In all variants, satellites locally train assigned expert paths and may compute layer-wise filter-importance statistics. The GS then derives pruning decisions from either aggregated global weights or client-side importance signals and centrally applies structured pruning by shrinking the corresponding expert paths. Satellite-informed methods transmit only compact per-layer importance vectors in addition to model updates; for a layer with $C_{\text{out}}$ filters, this overhead is $\mathcal{O}(C_{\text{out}})$ compared with $\mathcal{O}(K^2C_{\text{in}}C_{\text{out}})$ for convolution weights. Thus, pruning coordination incurs minimal communication overhead. While client-informed pruning better captures satellite-specific relevance under heterogeneous data, GS-side pruning provides consistent centralized control during aggregation~\cite{wu2023efficient}.

\noindent{\bf 1. Centralized weight-based pruning.}
After aggregation, the GS computes global filter scores $S_i$ on the aggregated expert model $\theta_t$ and removes a fixed fraction $r$ of the least important filters per layer. This baseline is simple but may ignore satellite-specific relevance under heterogeneous data distributions. Moreover, computing magnitude-based importance on aggregated weights can underestimate filter relevance when satellite parameters partially cancel after averaging. Formally, with $\bar{W}_i=\sum_k \alpha_k W_i^{(k)}$, GS-side scoring uses $S_i^{\text{GS}}=\lVert \bar{W}_i\rVert_1$, whereas satellite-informed voting can aggregate non-negative scores such as $S_i^{\text{vote}}=\sum_k \alpha_k \lVert W_i^{(k)}\rVert_1$, reducing sensitivity to sign cancellations.

\noindent{\bf 2. Satellite-informed \underline{\em fixed-ratio} pruning.}
Each participating satellite computes local filter scores and sends a vote vector $V_k$ to the GS. The GS aggregates votes by the mean, $V_{\text{agg}} = \frac{1}{K}\sum_{k=1}^{K} V_k$, and prunes a fixed fraction $r$ using $V_{\text{agg}}$. This incorporates local data signals while keeping a uniform pruning schedule.

\noindent{\bf 3. Satellite-informed \underline{\em adaptive} pruning.}
To avoid a fixed pruning ratio, the adaptive strategy determines the pruning strength for each layer based on the distribution of aggregated filter importance scores. Importance vectors are aggregated using a coordinate-wise median:
$
V_{\mathrm{agg}}=\mathrm{median}(V_1,\dots,V_K).
$
We analyze stabilized scores $s=\log(1+V_{\mathrm{agg}})$ and identify significant separations in the sorted scores. Let $s_{(1)}\leq\cdots\leq s_{(N)}$ denote the sorted scores and $g_i=s_{(i+1)}-s_{(i)}$ the consecutive gaps. A gap is considered significant if the maximum gap $g^*$ exceeds a robust MAD-based threshold:
\[
g^*>z_{\mathrm{thresh}}\hat{\sigma}, \qquad
\hat{\sigma}=1.4826\cdot\mathrm{MAD}(g),
\]
and satisfies a minimum relative separation criterion. Filters below the detected gap are pruned, while layers without clear separation are skipped. Safety constraints, including minimum channel retention and per-step pruning limits, prevent structural collapse.

Unlike target-based pruning, the proposed criterion prunes only layers with clear importance separation, making adaptive pruning a \emph{gap-constrained} rather than target-exact strategy. Therefore, we report both requested and achieved pruning ratios. Importantly, the current implementation focuses on structured convolutional channel pruning; extending pruning to fully connected expert heads requires coordinated reconstruction of dependent linear layers and is left for future work. Consequently, compression gains primarily reflect convolutional pruning, while retained heads continue to contribute to model footprint and runtime. This explains why modular pruning reduces peak memory and power on NVIDIA Jetson AGX Orin without necessarily improving batch-size-one latency, motivating the joint reporting of memory, latency, power, and parameter counts.

\section{Performance Evaluation}\label{sec:evaluation}


\noindent{\bf Dataset \& Architecture.} We simulate satellite constellations with non-overlapping data distributions, varying the number of clients as $S \in \{10,50,100\}$ and, for MoE variants, the number of experts as $E \in \{3,5,7,10\}$. We evaluate \textsc{COSMIC-FL} on six image-classification benchmarks: CIFAR-10~\cite{krizhevsky2009learning}, AID~\cite{xia2017aid}, EuroSAT~\cite{helber2019eurosat}, UCMerced~\cite{yang2010bag}, RESISC45~\cite{cheng2017remote}, and CIFAR-100~\cite{krizhevsky2009learning}. Dataset statistics and additional experimental details are provided in Supplementary Material~A.


\noindent{\bf Hyperparameters.} {Unless stated otherwise, }COSMIC-FL uses expert specialization for $T_E = 50$ rounds, gate finetuning for $T_G = 20$ rounds, and head finetuning for $T_H = 20$ rounds. For non-IID data heterogeneity we use Dirichlet concentration parameter $\alpha \in \{0.1, 0.3, 1.0, 10\}$. Batch size, architecture, learning rates, FedAvg/FedProx rounds, and remaining settings are given in Supplementary Material~B.

\noindent{\bf Pruning Protocol.} Pruning runs are initialized from the dense baseline; we apply four pruning rounds followed by six recovery rounds to reach the target pruning ratio. The full protocol is described in Supplementary Material~C.

\noindent{\bf Baselines.} We consider six baselines to evaluate the effectiveness of COSMIC-FL: FedAvg~\cite{mcmahan2017communication}, FedProx~\cite{li2020federated}, CentMoE~\cite{shazeer2017outrageously}, PM-MoE~\cite{feng2025pm}, and two resource-efficient FL methods, FedDropout~\cite{caldas2018expanding} and HeteroFL~\cite{diao2020heterofl}, where $r$ denotes the fraction of the full model trained by each client. We provide a brief description of each baseline in Supplementary Material~E.


{\noindent{\bf Metrics.} The server-side evaluation reports test accuracy, parameter count, analytical FLOPs, phase-specific trainable parameters, and communicated parameters. The embedded evaluation additionally reports mean and 95th-percentile inference latency, throughput, peak allocated CUDA memory, board power, and device temperature.}

The embedded inference benchmarking protocol (device, warm-up, timed batches, and measurement tooling) is detailed in Supplementary Material~G.

\begin{table}[!t]
\centering
\caption{Accuracy (\%) under IID settings with different numbers of satellites.}
\label{tab:fl_results_iid_clients}
\setlength{\tabcolsep}{4pt}
\resizebox{\linewidth}{!}{
\begin{tabular}{l|c|c|c|c|c}
\toprule
Method (\# Satellites) & CIFAR-10 & AID & EuroSAT & RESISC45 & UC Merced \\
\midrule

FedAvg (10)   & 88.50 & 79.39 & 97.41 & 83.73 & 83.81 \\
FedAvg (50)   & 82.03 & 62.88 & 94.48 & 72.78 & 56.51 \\
FedAvg (100)  & 75.46 & 53.24 & 91.07 & 60.40 & 49.21 \\
\midrule

FedProx (10)  & 86.99 & 78.80 & 97.30 & 83.02 & 83.17 \\
FedProx (50)  & 81.55 & 64.80 & 94.56 & 73.67 & 55.87 \\
FedProx (100) & 74.61 & 53.43 & 90.22 & 62.82 & 48.89 \\
\midrule

CentMoE (10)  & 91.08 & 83.09 & 98.44 & 87.36 & 87.30 \\
CentMoE (50)  & 82.56 & 68.63 & 95.15 & 67.56 & 75.24 \\
CentMoE (100) & 76.51 & 62.55 & 92.26 & 59.42 & 73.02 \\
\midrule

PM-MoE (10)   & 91.83 & 45.44 & 97.33 & 89.02 & 86.03 \\
PM-MoE (50)   & 76.57 & 70.67 & 95.93 & 70.53 & 73.97 \\
PM-MoE (100)  & 75.35 & 63.28 & 92.37 & 61.49 & 64.13 \\
\midrule

FedDropout ($r{=}0.75$, 10)  & 84.84 & 62.09 & 90.11 & 71.04 & 45.40 \\
FedDropout ($r{=}0.75$, 50)  & 69.25 & 33.22 & 79.67 & 42.81 & 18.41 \\
FedDropout ($r{=}0.75$, 100) & 56.20 & 23.31 & 69.96 & 28.35 & 14.60 \\
\midrule

HeteroFL ($r{=}0.75$, 10)  & 87.88 & 74.01 & 97.06 & 79.83 & 66.03 \\
HeteroFL ($r{=}0.75$, 50)  & 76.15 & 48.76 & 89.19 & 58.37 & 41.27 \\
HeteroFL ($r{=}0.75$, 100) & 65.35 & 37.51 & 78.76 & 43.93 & 31.75 \\
\midrule

{\bf COSMIC-FL } (Modular, 10)  & 76.37 & 76.75 & 92.67 & 79.62 & 70.16 \\
{\bf COSMIC-FL } (Modular, 50) & 69.42 & 58.26 & 88.15 & 66.02 & 50.16 \\
{\bf COSMIC-FL } (Modular, 100) & 63.16 & 52.05 & 82.19 & 53.84 & 38.41 \\
\midrule

{\bf COSMIC-FL } (Sliced, 10)  & 61.81 & 69.75 & 89.93 & 70.60 & 69.52 \\
{\bf COSMIC-FL } (Sliced, 50)   & 62.91 & 55.02 & 87.96 & 63.67 & 51.43 \\
{\bf COSMIC-FL } (Sliced, 100) & 65.76 & 46.37 & 83.15 & 51.04 & 41.27 \\

\bottomrule
\end{tabular}}
\end{table}

\begin{table}[!t]
\centering
\caption{Impact of data heterogeneity (Dirichlet-$\alpha$) on test accuracy (EuroSAT, 50 clients). Lower $\alpha$ = more non-i.i.d (see Supplementary Material~F for more results).}
\label{tab:ablation_noniid}
\setlength{\tabcolsep}{4pt}
\resizebox{\linewidth}{!}{
\begin{tabular}{l |*{4}{c}}
\toprule
Method & $\alpha{=}0.1$ & $\alpha{=}0.3$ & $\alpha{=}1.0$ & $\alpha{=}10.0$ \\
\midrule
FedAvg        & 81.33 & 90.89 & 93.74 & 94.41 \\
FedProx       & 83.00 & 90.63 & 93.37 & 94.48 \\
CentMoE       & 86.89 & 94.07 & 95.07 & 95.26 \\
PM-MoE        & 11.11 & 92.70 & 95.22 & 95.26 \\
FedDropout ($r{=}0.75$)    & 64.72 & 75.13 & 79.19 & 79.94 \\
HeteroFL ($r{=}0.75$)      & 74.98 & 81.70 & 86.17 & 89.54 \\
\midrule
{\bf COSMIC-FL (Modular)} & 67.26 & 82.37 & 86.67 & 88.37 \\
{\bf COSMIC-FL (Sliced)}  & 67.67 & 83.26 & 86.52 & 87.78 \\
\bottomrule
\end{tabular}}
\end{table}

\subsection{Experimental Results}

\subsubsection{Accuracy Comparison of \textsc{COSMIC-FL} vs.\ Baselines under i.i.d.\ Settings.}
Table~\ref{tab:fl_results_iid_clients} shows that accuracy generally declines as the number of satellites increases (10$\rightarrow$50$\rightarrow$100) and the datasets become more challenging (e.g., AID, RESISC45, and UC~Merced). CentMoE achieves the highest accuracy in most cases across all datasets due to full-model aggregation and larger capacity. 
In contrast, \textsc{COSMIC-FL} trades marginal accuracy loss for substantial reductions in phase-specific training and communication costs, which is critical for onboard learning under constrained compute, memory, energy, and contact windows. Compared with sub-model efficiency baselines at their strongest partial-capacity setting ($r{=}0.75$), \textsc{COSMIC-FL} consistently outperforms FedDropout across all remote-sensing benchmarks and constellation sizes, and surpasses HeteroFL on AID, RESISC45, and UC~Merced at 50 and 100 clients (e.g., RESISC45 with 50 clients: 66.02\%/63.67\% for Modular/Sliced vs.\ 58.37\%). HeteroFL remains competitive on CIFAR-10 and EuroSAT under smaller constellations.

\subsubsection{Accuracy Comparison of COSMIC-FL vs.\ Baselines (non-i.i.d) Settings.}Table~\ref{tab:ablation_noniid} reports test accuracy on EuroSAT under varying data heterogeneity (Dirichlet-$\alpha$, where smaller $\alpha$ indicates stronger non-i.i.d.\ distributions) with 50 clients. RESISC45 results are provided in Supplementary Material~F. On EuroSAT, all methods improve as $\alpha$ increases from 0.1 to 10. Under strong heterogeneity ($\alpha = 0.1$), \textsc{COSMIC-FL} (Modular) and \textsc{COSMIC-FL} (Sliced) achieve 67.26\% and 67.67\%, respectively, rising to about 86--88\% at $\alpha = 10$. FedAvg, FedProx, and CentMoE achieve about 81--87\% at $\alpha = 0.1$ and 94--95\% at $\alpha = 10$. PM-MoE drops to 11.11\% at $\alpha = 0.1$ but recovers to 92.70\% at $\alpha = 0.3$ and 95.26\% at $\alpha = 10$. Similar trends appear on RESISC45, indicating that \textsc{COSMIC-FL} follows the same heterogeneity-dependent behavior while prioritizing reduced staged training and communication cost.




\begin{figure*}[!t]
  \centering
  \includegraphics[width=\linewidth]{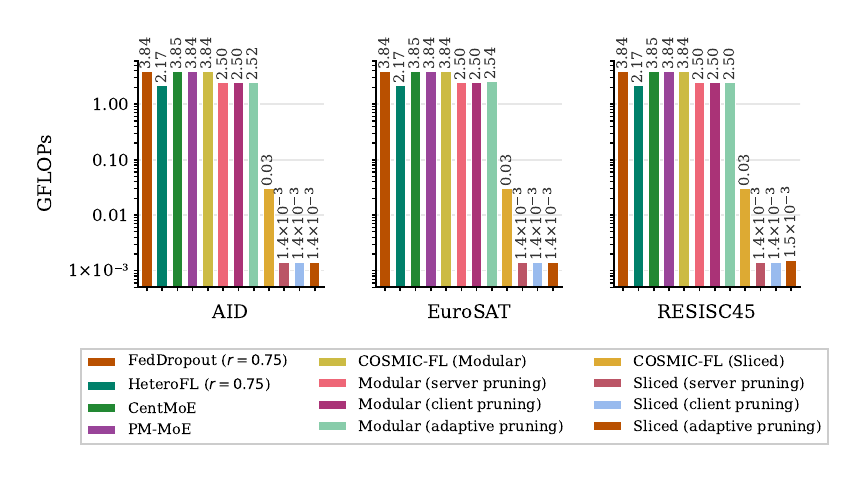}
  \caption{
    Inference cost (GFLOPs per sample) by method and dataset (AID, EuroSAT, RESISC45). Methods: CentMoE, COSMIC-FL (Modular), COSMIC-FL (Sliced), and their server-, client-, and adaptive-pruning variants.
  }
  \label{fig:gflops}
\end{figure*}

\subsubsection{Communication and Inference Efficiency of COSMIC-FL vs. Baselines.}
Fig.~\ref{fig:gflops} reports the inference cost of all methods. The complete communication analysis --- per-client per-round payloads and total uplink across all training phases --- is provided in Supplementary Material~F.

The analytical FLOP count captures arithmetic complexity but does not necessarily predict embedded latency. In particular, despite lower nominal FLOPs, the Sliced implementation can exhibit higher runtime on Jetson AGX Orin due to kernel execution overhead, tensor indexing, and memory behavior. Therefore, analytical FLOPs and measured embedded runtime should be viewed as complementary, rather than interchangeable, efficiency metrics.

\subsubsection{Sparse Expert Participation.}
Supplementary Material~F reports the full sparse expert participation study, where each client trains only experts with sufficient local samples instead of the full expert set. Under strong heterogeneity ($\alpha{=}0.1$) with 50 clients, sparse participation reduces the average active experts per client from 3.0 to 0.82 (Modular) and 0.92 (Sliced), reducing per-round uploads by 72.7\% and 69.3\%, respectively. Despite fewer updates, accuracy improves by 7.3\% (Modular) and 8.2\% (Sliced), as experts avoid noisy updates from clients lacking relevant classes. Rotation time also decreases by 26--29\%, with larger communication savings as the constellation scales (up to 89.3\% for Modular and 87.7\% for Sliced at 100 clients). Sparse participation maintains comparable or better accuracy than full expert training across scales. Unlike width-reduction methods such as FedDropout and HeteroFL, \textsc{COSMIC-FL} enables clients to skip entire experts based on local data coverage.

\begin{table}[!t]
\centering
\caption{Pruning strategy and ratio analysis for \textsc{COSMIC-FL} (Sliced and Modular) on EuroSAT with 50 satellites. Results report retained parameters (P, millions) and accuracy (\%). Additional results are provided in Supplementary Material~F.}
\label{tab:ablation_pruning_methods} \setlength{\tabcolsep}{4pt}
\begin{tabular}{c| c|cc|cc|cc}
\toprule \multirow{2}{*}{\rotatebox[origin=c]{90}{Model}} & \multirow{2}{*}{\rotatebox[origin=c]{90}{Prune}} & \multicolumn{2}{c}{Server} & \multicolumn{2}{c}{Client} & \multicolumn{2}{c}{Adaptive} \\ \cmidrule(lr){3-4} \cmidrule(lr){5-6} \cmidrule(lr){7-8} & & P & Acc & P & Acc & P & Acc \\
\midrule
\multirow{6}{*}{\rotatebox[origin=c]{90}{\makecell{COSMIC-FL\\(Sliced)}}} & 0 & 12.4 & 87.56 & 12.4 & 87.56 & 12.4 & 87.56 \\ & 10 & 11.5 & 83.74 & 11.5 & 82.07 & 11.5 & 83.15 \\ & 30 & 8.2 & 81.15 & 8.2 & 81.04 & 8.4 & 81.93 \\ & 50 & 5.6 & 80.67 & 5.6 & 82.04 & 5.8 & 82.22 \\ & 70 & 3.5 & 80.59 & 3.5 & 79.67 & 4.0 & 80.89 \\ & 80 & 2.7 & 79.63 & 2.7 & 75.41 & 3.0 & 78.70 \\
\midrule
\multirow{6}{*}{\rotatebox[origin=c]{90}{\makecell{COSMIC-FL\\(Modular)}}}& 0 & 19.8 & 87.37 & 19.8 & 87.37 & 19.8 & 87.37 \\ & 10 & 18.7 & 79.89 & 18.7 & 79.22 & 18.7 & 79.15 \\ & 30 & 16.4 & 80.11 & 16.4 & 79.78 & 16.6 & 80.41 \\ & 50 & 13.9 & 80.85 & 13.9 & 80.81 & 14.2 & 81.07 \\ & 70 & 11.2 & 81.00 & 11.2 & 81.63 & 11.3 & 81.33 \\ & 80 & 9.7 & 81.07 & 9.7 & 81.33 & 10.1 & 81.19 \\
\bottomrule \end{tabular}
\end{table}

\subsubsection{Effect of Structured Expert Pruning.}
Table~\ref{tab:ablation_pruning_methods} compares pruning strategies for \textsc{COSMIC-FL} on EuroSAT with 50 satellites; RESISC45 results are provided in Supplementary Material~F. All pruning methods substantially reduce parameters with modest accuracy loss. At the same pruning ratio, server-side, fixed-ratio client-side, and adaptive pruning achieve comparable performance, with larger differences observed for RESISC45 and Sliced-MoE. Adaptive pruning retains slightly more parameters due to its layer-wise gap-based decisions rather than strict ratio enforcement. The optimal pruning level is architecture-dependent: Modular-MoE remains competitive under aggressive pruning (70--80\%), while Sliced-MoE achieves its best trade-off under moderate pruning (30--50\%).

\subsubsection{Embedded Training Feasibility}
To assess deployment feasibility, we execute the proposed training pipeline on an NVIDIA Jetson AGX Orin with 64\,GB unified memory (Jetson Linux R36.4.7). This experiment validates embedded execution rather than serving as the primary training benchmark. For the sliced EuroSAT model, the pipeline achieves $87.78\%$ accuracy with a total runtime of $1{,}195.12$ minutes, including $1{,}103.19$ minutes of training, demonstrating the feasibility of staged training on embedded hardware.

Table~\ref{tab:jetson_inference_main} shows that FedAvg and FedProx provide the lowest-latency single-tower baselines. Among reduced-capacity methods ($r{=}0.75$), HeteroFL achieves the lowest peak memory, while FedDropout suffers larger accuracy degradation. PM-MoE achieves high accuracy but requires substantially more memory. Thus, the proposed pruned variants represent accuracy--resource trade-offs rather than universally faster models: Sliced pruning reduces memory relative to PM-MoE at the cost of accuracy. Additional profiling results are provided in Supplementary Material~G.

\begin{table}[t]
\centering
\caption{Jetson AGX Orin inference performance on EuroSAT and RESISC45. Reported metrics include classification accuracy (\%), latency (ms), peak allocated GPU memory (MB), and average board power (W).}
\label{tab:jetson_inference_main}
\setlength{\tabcolsep}{3pt}
\resizebox{\linewidth}{!}{
\begin{tabular}{l|rrrr|rrrr}
\toprule
&
\multicolumn{4}{c|}{EuroSAT} &
\multicolumn{4}{c}{RESISC45}\\
\cmidrule(lr){2-5}\cmidrule(l){6-9}
Method
& Acc & Lat. & Mem & Pow
& Acc & Lat. & Mem & Pow\\
\midrule
FedAvg         &94.3&3.66&38.2&4.90&72.1&3.66&38.3&4.96\\
FedProx        &94.6&3.65&38.2&4.99&72.1&3.66&38.3&5.04\\
FedDropout     &79.7&4.52&39.1&4.34&42.8&4.53&40.0&4.59\\
HeteroFL       &89.2&4.54&26.9&4.12&58.4&4.54&26.9&4.12\\
PM-MoE         &96.0&22.9&190.3&4.47&68.6&22.9&193.8&4.47\\
Modular        &79.1&11.2&61.5&4.53&58.9&11.2&55.0&4.52\\
Sliced         &79.5&21.4&42.3&4.07&43.7&21.5&44.5&4.06\\
\bottomrule
\end{tabular}}
\end{table}


\section{Conclusion}


This paper presented \textsc{COSMIC-FL}, a resource-aware FL framework for LEO satellite constellations that integrates staged expert training, complementary MoE architectures, semantic class-to-expert routing, and federated structured pruning. Extensive evaluations demonstrate an accuracy--efficiency trade-off across trained, communicated, and retained parameters, while embedded experiments on NVIDIA Jetson AGX Orin validate the feasibility of onboard execution. Further analysis shows that structured pruning reduces memory and power consumption, whereas runtime improvements depend on the expert architecture. These results highlight that FLOPs, parameter count, and embedded runtime provide complementary efficiency perspectives for onboard AI evaluation. Future work will investigate operator-level optimization of sliced execution and adaptive expert scheduling under dynamic satellite connectivity.



\bibliography{_References}

\clearpage
\appendix
\section{Dataset \& Architecture}
\subsection{Datasets}
\label{app:dataset}

We evaluate our framework on standard remote-sensing image classification benchmarks under federated learning using five widely-used datasets:
\begin{itemize}
  \item \textbf{AID}~\cite{xia2017aid}: 10{,}000 aerial scene images collected from Google Earth and annotated into 30 scene classes (e.g., airport, beach, desert); widely used for aerial scene classification.
  \item \textbf{EuroSAT}~\cite{helber2019eurosat}: 27{,}000 Sentinel-2 satellite image patches (64×64) covering 10 land-cover classes such as forest, crop, and water; commonly used for multispectral land-cover classification.
  \item \textbf{UCMerced}~\cite{yang2010bag}: 2{,}100 aerial images (256×256) distributed across 21 land-use classes; often used as a smaller benchmark for evaluating remote-sensing classification methods.
  \item \textbf{CIFAR-10}~\cite{krizhevsky2009learning}: 60{,}000 natural images (32×32) in 10 object classes; included as a standard vision benchmark to facilitate comparison with general federated learning and MoE methods.
  \item \textbf{RESISC45}~\cite{cheng2017remote}: 31{,}500 remote-sensing images (256×256) across 45 scene classes; provides a larger and more diverse benchmark for aerial scene classification.
  \item \textbf{CIFAR-100}~\cite{krizhevsky2009learning}: 60{,}000 natural images (32×32) in 100 object classes; included to evaluate scalability to larger label spaces.
\end{itemize}
For all datasets, we use a 70\%--15\%--15\% split for training, validation, and test. A 5\% portion of the training set is reserved for server-side gate pretraining. We simulate a constellation of satellites, each holding a non-overlapping portion of the data, and vary the number of satellites (clients) $S \in \{10, 50, 100\}$; for MoE variants we vary the number of experts $E \in \{3, 5, 7, 10\}$.

\subsection{Architecture}
The architecture uses a shared stem (1 conv block, 128 channels) and expert conv blocks (4 blocks, hidden dimension 256 --- per expert tower in Modular design, and as the total width partitioned into per-expert channel ranges in Sliced design), with a gating network and per-expert classification heads. The gating network takes the stem output, applies global average pooling, and passes the 128-d vector through a two-layer MLP (128 $\to$ 128 $\to$ $\mathit{num\_experts}$) with ReLU to produce expert logits. Each head is a three-layer MLP (256 $\to$ 1024 $\to$ 512 $\to$ $\mathit{num\_classes}$) with ReLU and dropout after the first two layers; there is one head per expert. For semantic class-to-path mapping we use a frozen ResNet18 (ImageNet-pretrained) to extract 512-d features, compute per-class mean vectors on the training set, and run k-means ($k = \mathit{num\_experts}$) to assign each class to an expert path. FedAvg, FedProx, and CentMoE use the same expert structure for fair comparison.

\section{Hyperparameters}
\label{app:hyperparameters}
We use batch size 64 and weight decay $10^{-4}$ for all methods. Head dropout is 0.3. For COSMIC-FL (sliced and modular), gate pretraining runs 40 epochs for sliced and 12 for modular, with learning rate for the stem $1.5 \times 10^{-4}$ and for the gate $2 \times 10^{-4}$. Expert specialization runs for $T_E = 50$ rounds, 5 local epochs, and learning rate $5 \times 10^{-4}$; gate finetuning uses $T_G = 20$ rounds and head finetuning $T_H = 20$ rounds, each with 4 local epochs and learning rates $2 \times 10^{-3}$ and $7 \times 10^{-4}$. FedAvg and FedProx use 50 rounds, 5 local epochs, and a learning rate $5 \times 10^{-4}$; FedProx uses the default proximal coefficient $\mu = 0.01$. For non-IID data heterogeneity, we use a Dirichlet concentration parameter $\alpha \in \{0.1, 0.3, 1.0, 10\}$. Studies that use different schedules state them explicitly: the sparse-participation study uses $T_E{=}15$ (Table~\ref{tab:sparse_expert_scales}), and the CIFAR-100 scalability study uses the two budgets given in Table~\ref{tab:ablation_experts_cifar100}.

\begin{table*}[!tp]
\centering
\caption{MAD-threshold sensitivity of adaptive pruning for the reported threshold sweep. Each cell reports achieved parameter pruning and test accuracy.}
\label{tab:mad_sensitivity_supp}
\resizebox{\textwidth}{!}{
\begin{tabular}{llrrrrrr}
\toprule
Dataset & Impl. & Target & $z{=}1$ & $z{=}2$ & $z{=}3$ & $z{=}10$ & $z{=}20$ \\
\midrule
EuroSAT & Modular & 30\% & 20.44 / 78.30 & 20.44 / 78.30 & 20.44 / 78.30 & 20.44 / 78.30 & 16.14 / 77.63 \\
EuroSAT & Modular & 50\% & 35.42 / 79.11 & 35.42 / 79.11 & 35.42 / 79.11 & 35.42 / 79.11 & 24.54 / 77.78 \\
EuroSAT & Modular & 70\% & 49.97 / 78.93 & 49.97 / 78.93 & 49.97 / 78.93 & 47.01 / 79.00 & 33.54 / 78.33 \\
\midrule
EuroSAT & Sliced & 30\% & 34.52 / 80.19 & 34.52 / 80.19 & 34.52 / 80.19 & 29.79 / 82.22 & 18.00 / 79.81 \\
EuroSAT & Sliced & 50\% & 53.30 / 79.52 & 53.30 / 79.52 & 53.30 / 79.52 & 46.47 / 81.63 & 29.94 / 77.37 \\
EuroSAT & Sliced & 70\% & 65.05 / 76.93 & 65.05 / 76.93 & 65.05 / 76.93 & 56.29 / 79.56 & 39.15 / 79.96 \\
\midrule
RESISC45 & Modular & 30\% & 23.70 / 57.31 & 23.70 / 57.31 & 23.70 / 57.31 & 23.70 / 57.31 & 22.28 / 57.27 \\
RESISC45 & Modular & 50\% & 39.81 / 58.87 & 39.81 / 58.87 & 39.81 / 58.87 & 39.81 / 58.87 & 30.81 / 58.18 \\
RESISC45 & Modular & 70\% & 55.11 / 59.78 & 55.11 / 59.78 & 55.11 / 59.78 & 49.32 / 59.29 & 44.45 / 58.80 \\
\midrule
RESISC45 & Sliced & 30\% & 34.14 / 43.78 & 34.14 / 43.78 & 34.14 / 43.78 & 27.01 / 45.44 & 16.08 / 43.31 \\
RESISC45 & Sliced & 50\% & 51.52 / 43.67 & 51.52 / 43.67 & 51.52 / 43.67 & 42.93 / 43.47 & 28.28 / 43.40 \\
RESISC45 & Sliced & 70\% & 65.03 / 37.67 & 65.03 / 37.67 & 65.03 / 37.67 & 60.24 / 35.87 & 38.10 / 44.24 \\
\bottomrule
\end{tabular}}
\end{table*}
\begin{table*}[!tp]
\centering
\caption{Impact of the number of experts on COSMIC-FL modular and sliced variants evaluated on EuroSAT and RESISC45 with 50 clients. Results report parameter counts (M) and classification accuracy (\%).}
\label{tab:ablation_experts}
\resizebox{\linewidth}{!}{
\begin{tabular}{l|cccc|cccc}
\toprule
Num. Experts & \multicolumn{4}{c}{EuroSAT} & \multicolumn{4}{c}{RESISC45} \\
\cmidrule(lr){2-5} \cmidrule(lr){6-9}
& \multicolumn{2}{c}{Modular} & \multicolumn{2}{c}{Sliced} & \multicolumn{2}{c}{Modular} & \multicolumn{2}{c}{Sliced} \\
\cmidrule(lr){2-3} \cmidrule(lr){4-5} \cmidrule(lr){6-7} \cmidrule(lr){8-9}
& Params (M) & Acc. (\%) & Params (M) & Acc. (\%) & Params (M) & Acc. (\%) & Params (M) & Acc. (\%) \\
\midrule
3 & 19.8 & 88.19 & 7.4 & 87.96 & 19.9 & 65.82 & 7.4 & 63.67 \\
5 & 33.0 & 73.56 & 9.0 & 73.85 & 33.2 & 55.56 & 9.0 & 40.27 \\
7 & 46.3 & 67.74 & 10.6 & 73.44 & 46.5 & 49.07 & 10.7 & 37.98 \\
Best & {\textbf{19.8}} & \textbf{88.19} & \textbf{7.4} & \textbf{87.96} & \textbf{19.9} & \textbf{65.82} & \textbf{7.4} & \textbf{63.67} \\
\bottomrule
\end{tabular}}
\end{table*}

\begin{table*}[!tp]
\centering
\caption{Impact of the number of experts on COSMIC-FL modular and sliced variants evaluated on CIFAR-100 (100 classes) under a short and a long training budget. Results report parameter counts (M) and classification accuracy (\%).}
\label{tab:ablation_experts_cifar100}
\resizebox{\linewidth}{!}{
\begin{tabular}{l|ccc|ccc}
\toprule
& \multicolumn{3}{c|}{Modular} & \multicolumn{3}{c}{Sliced} \\
\cmidrule(lr){2-4} \cmidrule(lr){5-7}
Num. Experts & Params (M) & Acc.$_{\text{short}}$ (\%) & Acc.$_{\text{long}}$ (\%) & Params (M) & Acc.$_{\text{short}}$ (\%) & Acc.$_{\text{long}}$ (\%) \\
\midrule
3  & 17.6 & \textbf{21.00} & \textbf{29.05} & 7.6  & \textbf{22.30} & \textbf{29.15} \\
5  & 29.3 & 18.37 & 25.21 & 9.2  & 15.46 & 19.31 \\
7  & 41.0 & 15.52 & 21.59 & 10.9 & 10.22 & 13.20 \\
10 & 58.6 & 14.03 & 19.96 & 13.4 &  9.13 & 12.72 \\
\bottomrule
\end{tabular}}
\end{table*}

\begin{table*}[!tp]
\centering
\caption{Test accuracy (\%) after each training stage of COSMIC-FL (Sliced) on EuroSAT with 50 clients under varying Dirichlet heterogeneity. $\delta$ = gain over the preceding stage.}
\label{tab:stage_progression}
\resizebox{\linewidth}{!}{
\begin{tabular}{l|cc|cc|cc|cc}
\toprule
& \multicolumn{2}{c|}{$\alpha{=}0.1$} & \multicolumn{2}{c|}{$\alpha{=}0.3$} & \multicolumn{2}{c|}{$\alpha{=}1.0$} & \multicolumn{2}{c}{$\alpha{=}10$} \\
Stage & Acc. & $\delta$ & Acc. & $\delta$ & Acc. & $\delta$ & Acc. & $\delta$ \\
\midrule
Gate pretraining (init.)   & 8.07  & --     & 8.22  & --     & 8.52  & --     & 8.48  & --     \\
+ Expert specialization    & 53.15 & +45.08 & 73.48 & +65.26 & 75.85 & +67.33 & 76.37 & +67.89 \\
+ Gate fine-tuning         & 57.04 & +3.89  & 76.52 & +3.04  & 80.19 & +4.34  & 80.07 & +3.70  \\
+ Head fine-tuning         & \textbf{67.67} & +10.63 & \textbf{83.26} & +6.74 & \textbf{86.52} & +6.33 & \textbf{87.78} & +7.71 \\
\bottomrule
\end{tabular}
}
\end{table*}

\subsection{Simulation Environment}
\label{sec:env}
Server-side federated training simulations were run on a workstation with an NVIDIA GeForce RTX~4090 GPU (24\,GB), an Intel Core i9-14900K CPU, and 125\,GB RAM, running Ubuntu 24.04 with Python 3.10, PyTorch 2.5.1, and CUDA 12.1. All experiments use a fixed random seed (42) that controls data partitioning, model initialization, and client sampling; the seeding utilities are included in the released code. Unless stated otherwise, all reported training results are obtained from a single run under this fixed seed, while embedded measurements are averaged over two independent runs. Embedded measurements use the NVIDIA Jetson AGX Orin platform described in Supplementary Material~G.

\begin{figure*}[!tp]
  \centering
  \begin{subfigure}[t]{0.33\textwidth}
    \centering
    \includegraphics[width=\linewidth]{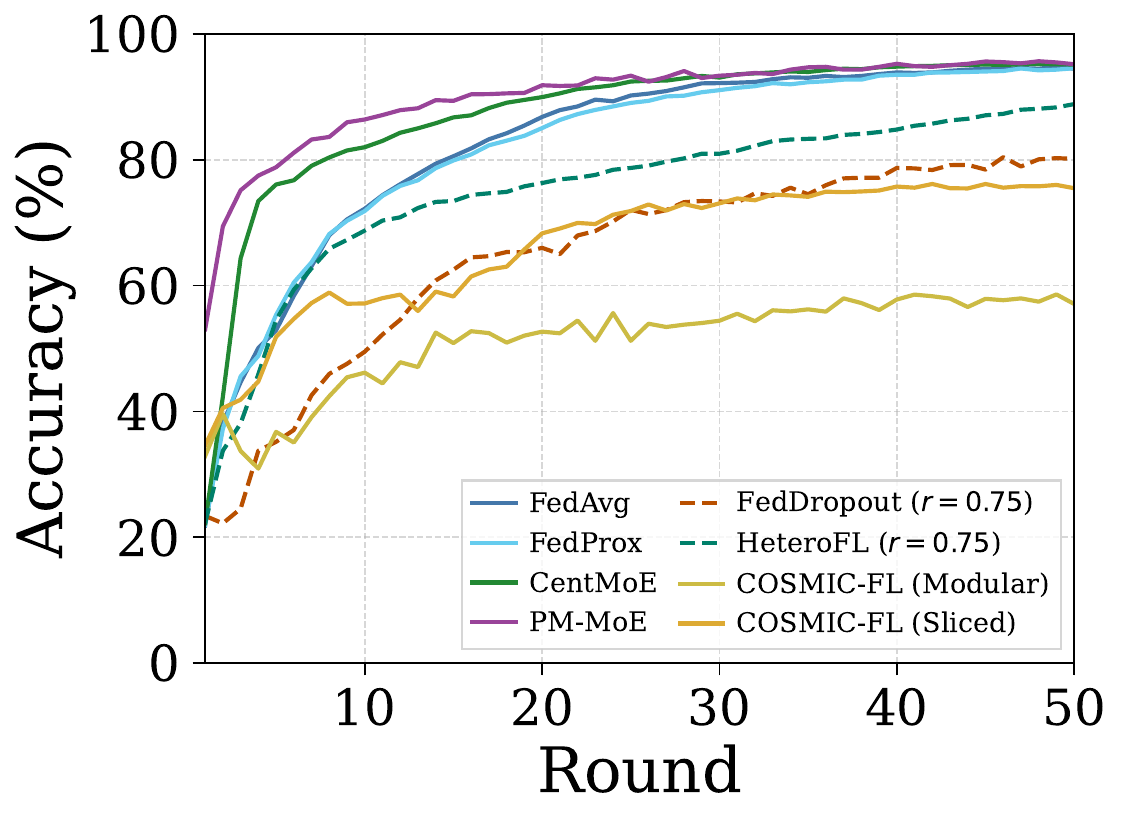}
    \caption{EuroSAT}
  \end{subfigure}
  \hfill
  \begin{subfigure}[t]{0.33\textwidth}
    \centering
    \includegraphics[width=\linewidth]{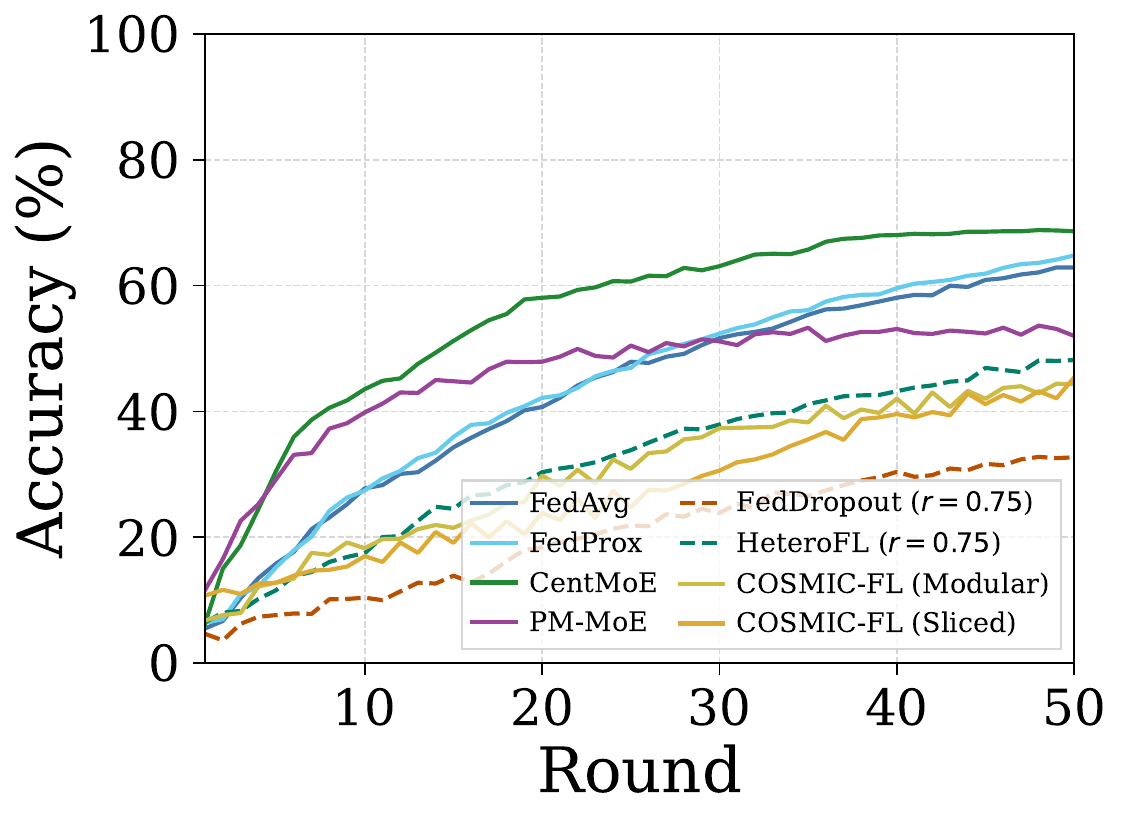}
    \caption{AID}
  \end{subfigure}
  \hfill
  \begin{subfigure}[t]{0.33\textwidth}
    \centering
    \includegraphics[width=\linewidth]{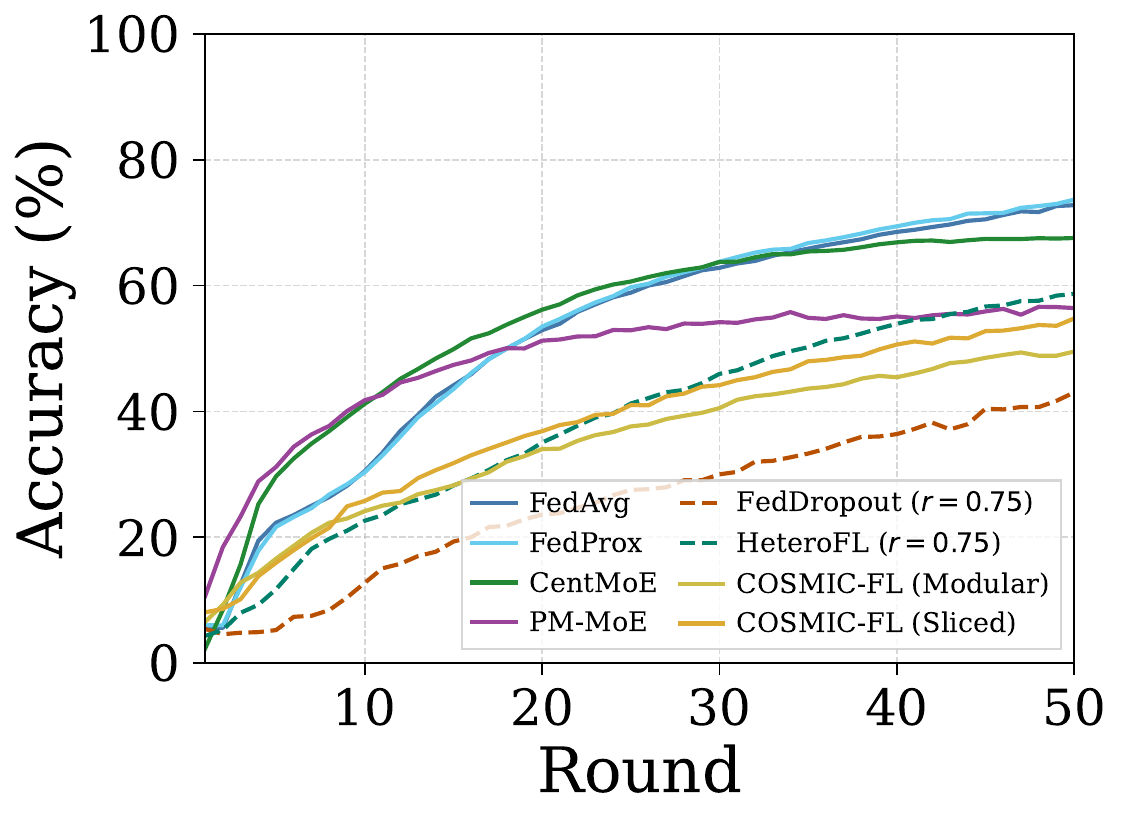}
    \caption{RESISC45}
  \end{subfigure}

  \caption{Test accuracy during the expert-specialization phase with 50 rounds and 50 clients on EuroSAT, AID, and RESISC45. Curves compare FedAvg, FedProx, CentMoE, PM-MoE, and the COSMIC-FL variants.}
  \label{fig:accuracy_expert}
\end{figure*}
\section{Pruning Protocol}
\label{app:pruning}
For the pruning ablation, we initialize all pruning runs from the corresponding dense baseline obtained in the preceding experiment. The dense model continues to be trained for 10 additional epochs, which serves as the reference model for the pruning-stage reported as ``No Pruning'' in the pruning ablation table of the main paper. For each pruning strategy, pruning is applied over four pruning rounds followed by six recovery rounds. The per-round pruning rate is chosen such that the desired final pruning ratio (10\%, 30\%, 50\%, 70\%, or 80\%) is reached at the end of the pruning schedule.

\section{MAD Threshold Sensitivity}
\label{app:mad_sensitivity}

To analyze the robustness of the adaptive pruning criterion, we evaluate the MAD-based gap threshold $z_{\mathrm{thresh}}$ (Table~\ref{tab:mad_sensitivity_supp}) for requested pruning targets of $30\%$, $50\%$, and $70\%$ on EuroSAT and RESISC45 using both the modular and sliced implementations. The adaptive method is gap-constrained rather than target-exact: it prunes a layer only when the sorted importance scores exhibit a sufficiently pronounced separation. Consequently, the achieved pruning ratio may differ from the requested pruning target.

Low and moderate thresholds often lead to identical pruning decisions, indicating that the detected score gaps are sufficiently pronounced. Higher thresholds make the adaptive rule more conservative and reduce the achieved pruning ratio. This behavior supports interpreting the requested pruning target as an upper-level compression objective, while the achieved pruning ratio reflects the layer-wise compressibility detected by the MAD-based gap test.

\section{Baseline descriptions}
\label{app:baselines}
{FedAvg}~\cite{mcmahan2017communication} performs standard averaging of client updates. {FedProx}~\cite{li2020federated} adds a proximal term to the local objective to limit client drift. {CentMoE}~\cite{shazeer2017outrageously} trains a central mixture-of-experts model with federated aggregation. {PM-MOE}~\cite{feng2025pm} first trains a FedPer-style shared body with personalized heads, then fine-tunes a mixture-of-experts over those heads so clients can combine converged personalized parameters from others. {FedDropout}~\cite{caldas2018expanding} reduces client cost by having each client train a randomly selected sub-model of the global network; the server aggregates the sparse updates back into the full model. {HeteroFL}~\cite{diao2020heterofl} assigns clients width-scaled sub-models of a shared global architecture and aggregates overlapping parameters, allowing heterogeneous client capacities. For both methods, $r$ denotes the fraction of the full model trained by each client ($r{=}1.0$ recovers full-model training); we evaluate $r \in \{0.25, 0.5, 0.75, 1.0\}$ to match the resource-constrained satellite setting.

\section{More Results}
\label{app:more_results}

\subsection{Effect of Number of Experts}
Tables~\ref{tab:ablation_experts} and~\ref{tab:ablation_experts_cifar100} show that using more experts does not necessarily improve performance. Across datasets spanning a 10$\times$ range of class counts --- EuroSAT (10 classes), RESISC45 (45), and CIFAR-100 (100) --- the best accuracy is always achieved with $E{=}3$, the smallest configuration, and accuracy declines monotonically as $E$ grows (on CIFAR-100: 29.1\% $\rightarrow$ 25.2\% $\rightarrow$ 21.6\% $\rightarrow$ 20.0\% for Modular at $E{=}3,5,7,10$). The CIFAR-100 study (Table~\ref{tab:ablation_experts_cifar100}) uses 50 clients, IID partitioning, and the semantic class-to-expert mapping, and reports two training budgets: short ($T_E{=}15$, $T_G{=}20$, $T_H{=}20$) and long ($T_E{=}30$, $T_G{=}20$, $T_H{=}40$). For Modular, increasing to five or seven experts reduces accuracy (e.g., EuroSAT: 88.19\% $\rightarrow$ 73.56\% $\rightarrow$ 67.74\%; RESISC45: 65.82\% $\rightarrow$ 55.56\% $\rightarrow$ 49.07\%) while parameters grow from 19.8M to 33.0M to 46.3M on EuroSAT and from 19.9M to 33.2M to 46.5M on RESISC45. For Sliced, three experts also yields the highest accuracy (87.96\% on EuroSAT, 63.67\% on RESISC45); with five or seven experts, accuracy drops, and on RESISC45 the drop is large (to 40.27\% and 37.98\%). A likely explanation is that, under fixed communication and training budgets, a larger expert pool makes expert specialization harder to stabilize and weakens the effective signal per branch--- each expert sees fewer samples, consistent with observations in centralized MoE that the useful number of experts tracks data volume rather than class count~\cite{videau2024mixture}, and with large-scale vision MoE models that serve 18k--21k classes with a fixed 32 experts per layer~\cite{riquelme2021scaling}.

These results directly address scalability to larger label spaces: the number of experts $E$ is decoupled from the number of classes $C$. Scaling from 10 to 100 classes does not require growing $E$ --- instead, the semantic class-to-expert mapping assigns a larger cluster of classes to each expert, and $E{=}3$ remains optimal across the entire range. The scaling knob is therefore \emph{classes-per-expert}, not experts-per-class, mirroring how federated MoE methods operate with $E \ll C$~\cite{dun2023fedjets}. The ranking is also invariant to the training budget: doubling the rotation rounds improves all configurations but widens the advantage of $E{=}3$ (the Modular $E{=}3$ vs.\ $E{=}10$ gap grows from 7.0 to 9.1 points, and the Sliced gap from 13.2 to 16.4), ruling out the possibility that larger expert pools merely converge more slowly. We note that absolute accuracy on CIFAR-100 is bounded by the compact onboard model capacity; the relevant evidence is the trend in $E$, which is consistent at every class count.

\subsubsection{Accuracy over Expert-Specialization Rounds}

Figure~\ref{fig:accuracy_expert} summarizes the evolution of test accuracy during the expert-specialization phase across all three datasets.
Across all three datasets (EuroSAT, AID, RESISC45), test accuracy increases steadily over the 50 expert-specialization rounds, with CentMoE and the single-tower baselines (FedAvg, FedProx) converging the fastest and to the highest accuracy. FedProx typically provides a small but consistent improvement over FedAvg, especially on RESISC45. PM-MoE starts lower and converges more slowly than CentMoE, but eventually reaches competitive accuracy on EuroSAT and AID. Our COSMIC-FL variants (Modular and Sliced) also start from lower accuracy and converge more slowly, reflecting their lower-capacity, modular design and the fact that only parts of the model are updated in this phase. Overall, CentMoE and the single-tower baselines converge fastest and to the highest accuracy, while PM-MoE and COSMIC-FL trade some accuracy for different personalization and efficiency properties.

\begin{table}[!t]
\centering
\caption{Sparse expert participation vs.\ full (normal) expert training for \textsc{COSMIC-FL} (EuroSAT, Dirichlet $\alpha{=}0.1$, 3 experts, 15 rotation rounds) across 10, 50, and 100 clients. With sparse participation, a client only trains experts for which it holds sufficient local samples, so it does not have to train the whole expert set each round. As the constellation scales, each client holds fewer samples, so fewer experts pass the ownership threshold: the communication saving grows (40\%$\to$89\% for Modular, 43\%$\to$88\% for Sliced) while validation accuracy is maintained or improved.}
\label{tab:sparse_expert_scales}
\resizebox{\linewidth}{!}{
\begin{tabular}{l|cc|cc}
\toprule
& \multicolumn{2}{c|}{Modular} & \multicolumn{2}{c}{Sliced} \\
Metric & Sparse & Normal & Sparse & Normal \\
\midrule
\multicolumn{5}{c}{\textit{10 clients}} \\
\midrule
Val.\ accuracy (\%)                     & \textbf{76.22} & 75.78 & \textbf{80.41} & 78.85 \\
Expert participation rate (\%)          & 60.0 & 100   & 56.7 & 100  \\
Experts trained / client / round        & 1.80 & 3.00  & 1.70 & 3.00 \\
Params transferred / client / round (M) & 11.89 & 19.81 & 4.19 & 7.39 \\
Total params transferred (G)            & 1.78 & 2.97  & 0.63 & 1.11 \\
Communication saving (\%)               & 40.0 & --    & 43.3 & --   \\
Rotation wall time (s)                  & 784  & 834   & 836  & 887  \\
\midrule
\multicolumn{5}{c}{\textit{50 clients}} \\
\midrule
Val.\ accuracy (\%)                     & \textbf{58.69} & 51.39 & \textbf{66.56} & 58.41 \\
Expert participation rate (\%)          & 27.3 & 100   & 30.7 & 100  \\
Experts trained / client / round        & 0.82 & 3.00  & 0.92 & 3.00 \\
Params transferred / client / round (M) & 5.42 & 19.81 & 2.27 & 7.39 \\
Total params transferred (G)            & 4.06 & 14.86 & 1.70 & 5.54 \\
Communication saving (\%)               & 72.7 & --    & 69.3 & --   \\
Rotation wall time (s)                  & 658  & 924   & 718  & 971  \\
\midrule
\multicolumn{5}{c}{\textit{100 clients}} \\
\midrule
Val.\ accuracy (\%)                     & \textbf{59.87} & 44.63 & \textbf{53.52} & 52.67 \\
Expert participation rate (\%)          & 10.7 & 100   & 12.3 & 100  \\
Experts trained / client / round        & 0.32 & 3.00  & 0.37 & 3.00 \\
Params transferred / client / round (M) & 2.11 & 19.81 & 0.92 & 7.39 \\
Total params transferred (G)            & 3.17 & 29.72 & 1.37 & 11.09 \\
Communication saving (\%)               & 89.3 & --    & 87.7 & --   \\
Rotation wall time (s)                  & 558  & 1015  & 610  & 1071 \\
\bottomrule
\end{tabular}}
\end{table}

\subsection{Contribution of Each Training Stage}
Table~\ref{tab:stage_progression} decomposes the final \emph{Sliced} EuroSAT accuracies of the non-i.i.d.\ comparison in the main paper into the contributions of the individual training stages: the runs are identical to those reported there (50 clients, semantic class-to-expert mapping, reference schedule $T_E{=}50$, $T_G{=}20$, $T_H{=}20$), so the last row of the progression reproduces the corresponding main-paper entries exactly. The initialization row is measured after server-side gate pretraining, where the shared stem and gate are trained but all experts are still randomly initialized; accuracy is accordingly near chance. Three observations follow. First, every stage contributes positively at every heterogeneity level. Second, expert specialization provides by far the largest gain (+45 to +68 points), growing with $\alpha$ as local distributions approach i.i.d. Third, the fine-tuning stages divide the remaining gap: gate fine-tuning adds a consistent +3.0--4.3 points while aligning routing accuracy to 75--83\%, and head fine-tuning contributes most under strong heterogeneity (+10.6 points at $\alpha{=}0.1$ versus +6.3--7.7 elsewhere), consistent with the heads consolidating knowledge across experts precisely when local label distributions are most skewed. We note these are sequential incremental gains under a fixed schedule rather than a leave-one-out ablation. The same qualitative pattern holds for the Modular variant at $\alpha{=}0.1$ under the shorter sparse-study schedule ($T_E{=}15$): 17.41 $\rightarrow$ 39.22 $\rightarrow$ 43.56 $\rightarrow$ 50.89.

\begin{table*}[!tp]
\centering
\caption{Impact of data heterogeneity (Dirichlet-$\alpha$) on test accuracy (EuroSAT \& RESISC45, 50 clients). Lower $\alpha$ = more non-i.i.d.}
\label{tab:ablation_noniid_full}
\resizebox{\linewidth}{!}{
\begin{tabular}{l |*{4}{c}| *{4}{c}}
\toprule
Method & \multicolumn{4}{c|}{EuroSAT} & \multicolumn{4}{c}{RESISC45} \\
\cmidrule(lr){2-5} \cmidrule(lr){6-9}
& $\alpha{=}0.1$ & $\alpha{=}0.3$ & $\alpha{=}1.0$ & $\alpha{=}10.0$
& $\alpha{=}0.1$ & $\alpha{=}0.3$ & $\alpha{=}1.0$ & $\alpha{=}10.0$ \\
\midrule
FedAvg        & 81.33 & 90.89 & 93.74 & 94.41 & 57.69 & 66.64 & 69.24 & 72.2 \\
FedProx       & 83.00 & 90.63 & 93.37 & 94.48 & 62.27 & 68.42 & 70.98 & 73.27 \\
CentMoE       & 86.89 & 94.07 & 95.07 & 95.26 & 61.49 & 66.60 & 66.96 & 67.36 \\
PM-MoE       & 11.11 & 92.70 & 95.22 & 95.26 & 29.91 & 45.44 & 65.91 & 71.56 \\

FedDropout ($r{=}0.75$)    & 64.72 & 75.13 & 79.19 & 79.94 & 33.28 & 36.84 & 41.09 & 43.73 \\
HeteroFL ($r{=}0.75$)      & 74.98  & 81.70  & 86.17  & 89.54  & 49.65  & 55.75  & 58.32  & 59.14  \\

\midrule
{\bf COSMIC-FL (Modular)} & 67.26 & 82.37 & 86.67 & 88.37 & 47.00 & 57.02 & 62.38 & 65.58 \\
{\bf  COSMIC-FL (Sliced)}  & 67.67 & 83.26 & 86.52 & 87.78 & 46.49 & 57.11 & 61.51 & 63.18 \\
\bottomrule
\end{tabular}}
\end{table*}

\begin{figure*}[!tp]
  \centering
  \begin{subfigure}[t]{0.32\textwidth}
    \centering
    \includegraphics[width=\linewidth]{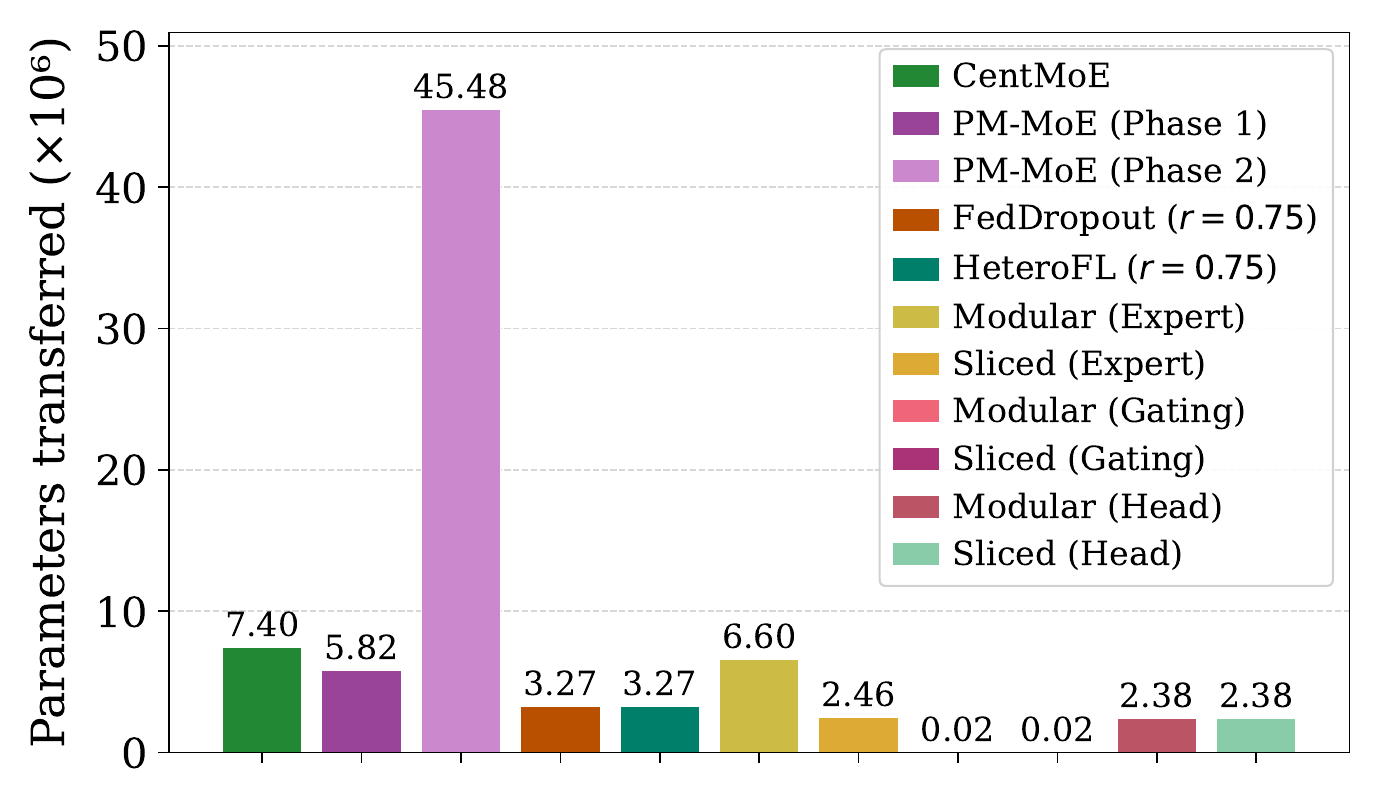}
    \caption{EuroSAT}
  \end{subfigure}
  \hfill
  \begin{subfigure}[t]{0.32\textwidth}
    \centering
    \includegraphics[width=\linewidth]{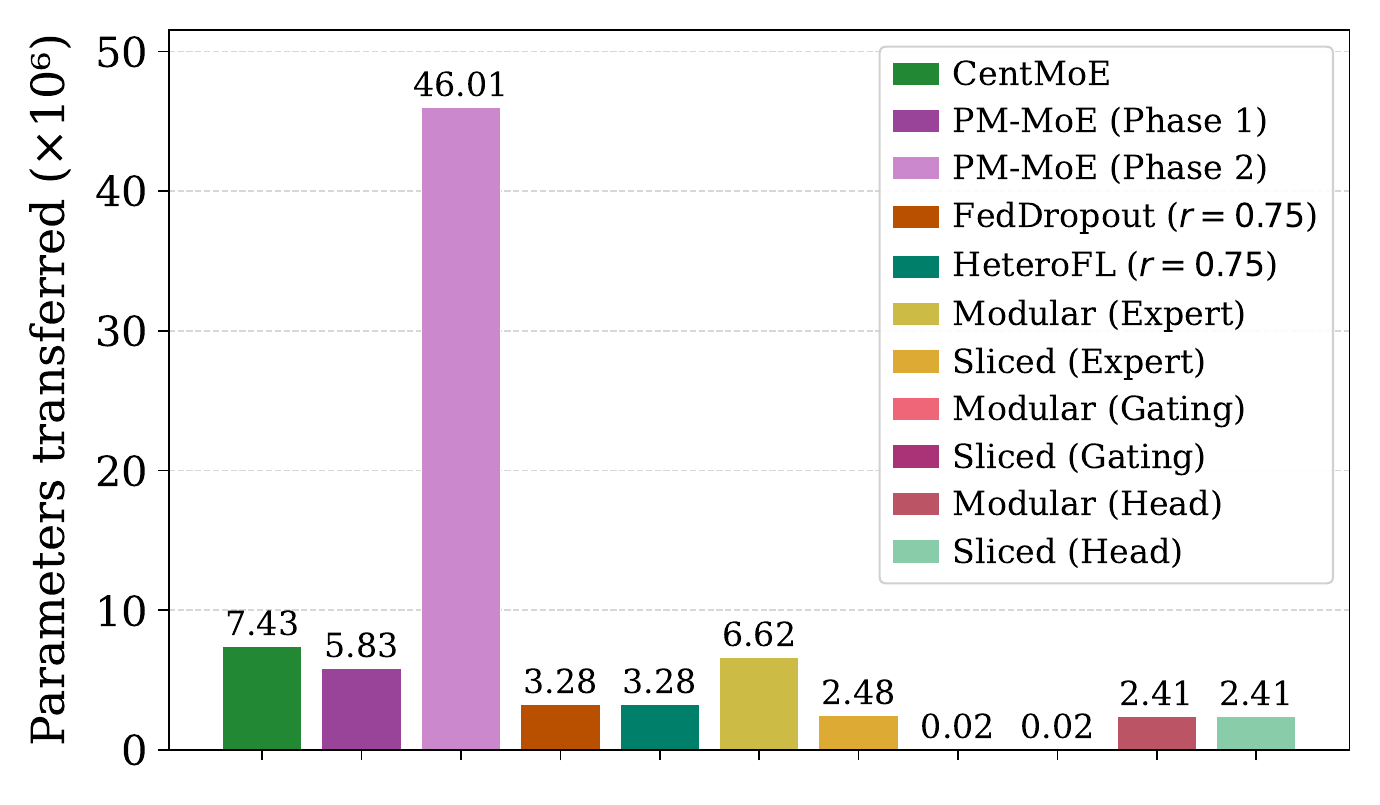}
    \caption{AID}
  \end{subfigure}
  \hfill
  \begin{subfigure}[t]{0.32\textwidth}
    \centering
    \includegraphics[width=\linewidth]{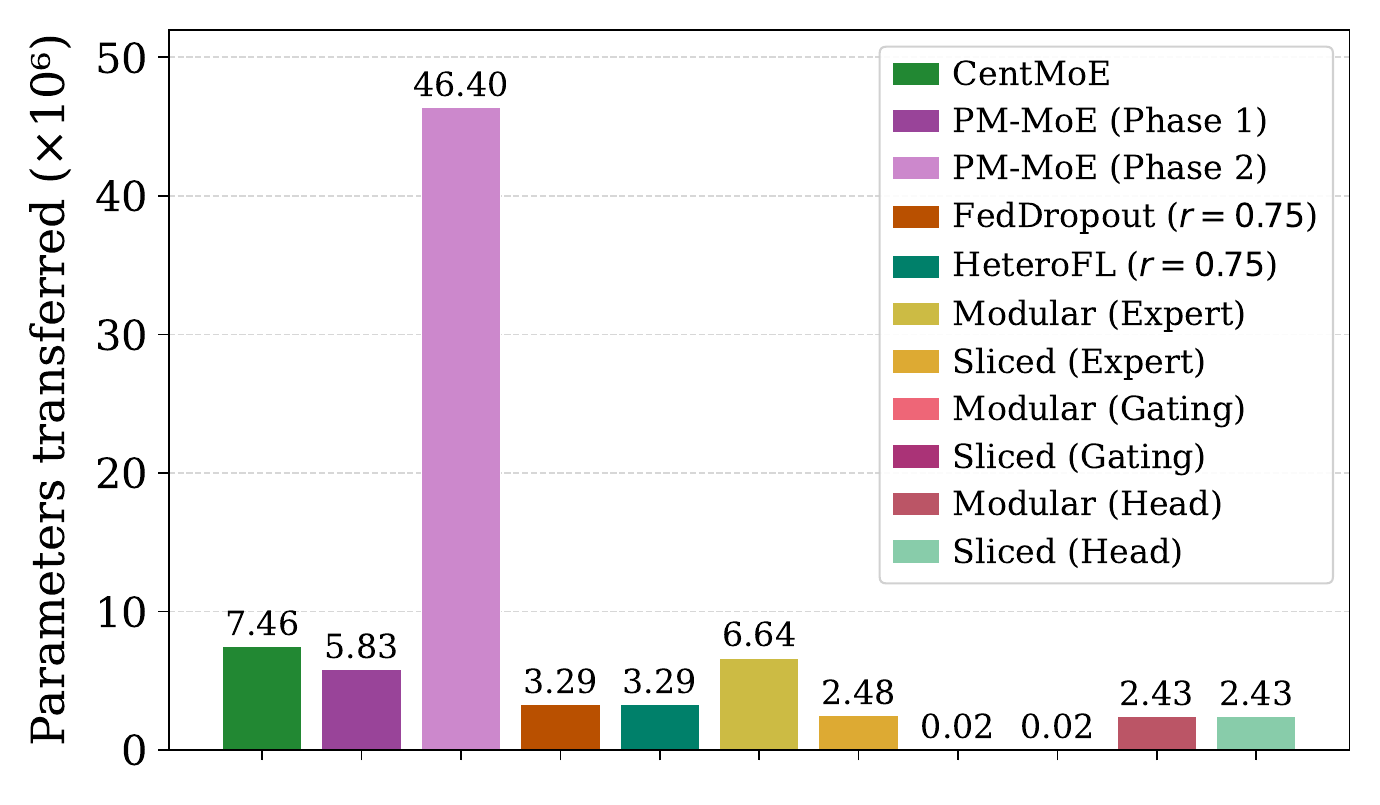}
    \caption{RESISC45}
  \end{subfigure}

  \caption{Per-client parameters trained and uploaded per round (millions) on EuroSAT, AID, and RESISC45. Each phase communicates only the parameters being optimized. CentMoE and PM-MoE train full models, while COSMIC-FL (Sliced/Modular) separates expert, gating, and head components.}
  \label{fig:params_transferred_supp}
\end{figure*}

\begin{table*}[!tp]
\centering
\caption{Total uplink communication across all training phases on EuroSAT, computed from measured per-client per-round payloads under the reference schedule ($T_E{=}50$, $T_G{=}20$, $T_H{=}20$). Sparse-expert variants scale rotation communication by the measured expert-participation rate $\rho$ (Table~\ref{tab:sparse_expert_scales}). ``--'' denotes methods without separate gate or head phases. $^{\dagger}$PM-MoE communication is derived from the implementation, where Phase~2 aggregates the full model including body, heads, and gate, causing client-dependent scaling.}
\label{tab:total_comm}
\resizebox{\linewidth}{!}{
\begin{tabular}{l|ccc|cc}
\toprule
& \multicolumn{3}{c|}{Payload / client / round (M)} & \multicolumn{2}{c}{Total uplink (G)} \\
Method & Rotation & Gate & Head & 50 clients & 100 clients \\
\midrule
FedAvg (full model)        & 5.82  & --   & --   & 14.5 & 29.1 \\
FedProx (full model)       & 5.82  & --   & --   & 14.5 & 29.1 \\
CentMoE                    & 7.40  & --   & --   & 18.5 & 37.0 \\
PM-MoE$^{\dagger}$         & 5.02  & --   & --   & 58.0 & 195.4 \\
FedDropout ($r{=}0.75$)    & 3.27  & --   & --   & 8.2  & 16.4 \\
HeteroFL ($r{=}0.75$)      & 3.27  & --   & --   & 8.2  & 16.4 \\
COSMIC-FL Modular          & 19.81 & 0.02 & 2.38 & 51.9 & 103.9 \\
COSMIC-FL Sliced           & 7.43  & 0.02 & 2.38 & 21.0 & 41.9 \\
COSMIC-FL Modular (sparse expert training) & $19.81\,\rho$ & 0.02 & 2.38 & \textbf{15.9} & \textbf{15.4} \\
COSMIC-FL Sliced (sparse expert training)  & $7.43\,\rho$  & 0.02 & 2.38 & \textbf{8.1}  & \textbf{9.4} \\
\bottomrule
\end{tabular}}
\end{table*}

\subsection{Sparse Expert Participation across Constellation Sizes}
Table~\ref{tab:sparse_expert_scales} reports the sparse expert participation study discussed in the main paper across 10, 50, and 100 clients. As the constellation scales, each client holds fewer samples, so fewer experts pass the ownership threshold: the communication saving grows from 40.0\% to 89.3\% (Modular) and from 43.3\% to 87.7\% (Sliced), while sparse validation accuracy remains at or above full training at every scale. Under strong heterogeneity ($\alpha{=}0.1$) with 50 clients, sparse participation reduces the average active experts per client from 3.0 to 0.82 (Modular) and 0.92 (Sliced), reducing per-round uploads by 72.7\% and 69.3\%, respectively. Despite fewer updates, accuracy improves by 7.3\% (Modular) and 8.2\% (Sliced), as experts avoid noisy updates from clients lacking relevant classes, and rotation wall time decreases by 26--29\%. Unlike width-reduction methods such as FedDropout and HeteroFL, which shrink every layer but still require every client to train in every round, \textsc{COSMIC-FL} enables clients to skip entire experts based on local data coverage.

\subsection{Data Heterogeneity on EuroSAT and RESISC45}
Table~\ref{tab:ablation_noniid_full} reports the full data-heterogeneity study on both EuroSAT and RESISC45 (the main paper presents the EuroSAT columns). RESISC45 follows the same trends: all methods improve with larger $\alpha$, PM-MoE collapses under strong heterogeneity ($\alpha{=}0.1$), and the COSMIC-FL variants track the heterogeneity-dependent behavior of the full-model baselines at a fraction of their staged training and communication cost.

\subsection{Per-Round Communication Across Training Phases}
Fig.~\ref{fig:params_transferred_supp} reports the per-client per-round communication analysis on EuroSAT, AID, and RESISC45. CentMoE trains approximately 7.4--7.5M parameters per round, while \textsc{COSMIC-FL} reduces this to 6.6M for Modular-MoE and 2.5M for Sliced-MoE during expert specialization. Considering all expert, head, and gate updates, Sliced-MoE transfers 4.9M parameters per round, reducing communication by 34\% compared with CentMoE and 46\% compared with Modular-MoE. Gate fine-tuning introduces negligible overhead ($\sim$0.02M). The phase-wise pattern is consistent across datasets: Sliced-MoE achieves superior parameter efficiency through shared representations, whereas Modular-MoE trades higher communication cost for greater expert capacity and flexibility.

\begin{table*}[!t]
\centering
\caption{Full capacity-fraction sweep for FedDropout and HeteroFL: accuracy (\%) under IID settings with different numbers of satellites.}
\label{tab:submodel_r_sweep}
\resizebox{\linewidth}{!}{
\begin{tabular}{l|c|c|c|c|c}
\toprule
Method (\# Satellites) & CIFAR-10 & AID & EuroSAT & RESISC45 & UC Merced \\
\midrule
FedDropout ($r{=}0.25$, 10)  & 31.53 & 14.33 &  33.72 & 12.42 &  5.08 \\
FedDropout ($r{=}0.25$, 50)  & 21.75 &  6.30 &  19.59 &  6.62 &  7.62 \\
FedDropout ($r{=}0.25$, 100) & 14.41 &  5.56 &  18.24 &  4.47 &  4.76 \\
FedDropout ($r{=}0.5$, 10)   & 65.01 & 37.44 & 66.33 & 35.04 & 13.65 \\
FedDropout ($r{=}0.5$, 50)   & 43.92 & 13.46 & 59.98 & 16.42 &  9.52 \\
FedDropout ($r{=}0.5$, 100)  & 38.71 &  6.90 & 47.13 &  9.41 &  8.57 \\
FedDropout ($r{=}0.75$, 10)  & 84.84 & 62.09 & 90.11 & 71.04 & 45.40 \\
FedDropout ($r{=}0.75$, 50)  & 69.25 & 33.22 & 79.67 & 42.81 & 18.41 \\
FedDropout ($r{=}0.75$, 100) & 56.20 & 23.31 & 69.96 & 28.35 & 14.60 \\
FedDropout ($r{=}1.0$, 10)   & 88.80 & 77.23 & 97.44 & 81.01 & 73.02 \\
FedDropout ($r{=}1.0$, 50)   & 79.89 & 55.19 & 91.26 & 64.77 & 45.08 \\
FedDropout ($r{=}1.0$, 100)  & 70.25 & 43.94 & 82.65 & 51.36 & 31.43 \\
\midrule
HeteroFL ($r{=}0.25$, 10)  & 79.57 & 58.61 & 92.91 & 63.53 & 45.08 \\
HeteroFL ($r{=}0.25$, 50)  & 57.45 & 32.08 & 76.13 & 33.78 & 20.95 \\
HeteroFL ($r{=}0.25$, 100) & 46.95 & 22.44 & 68.35 & 24.57 & 17.46 \\
HeteroFL ($r{=}0.5$, 10)   & 86.41 & 69.86 & 96.33 & 75.09 & 60.63 \\
HeteroFL ($r{=}0.5$, 50)   & 69.04 & 44.27 & 82.54 & 49.31 & 30.48 \\
HeteroFL ($r{=}0.5$, 100)  & 58.93 & 32.95 & 75.31 & 36.17 & 25.40 \\
HeteroFL ($r{=}0.75$, 10)  & 87.88 & 74.01 & 97.06 & 79.83 & 66.03 \\
HeteroFL ($r{=}0.75$, 50)  & 76.15 & 48.76 & 89.19 & 58.37 & 41.27 \\
HeteroFL ($r{=}0.75$, 100) & 65.35 & 37.51 & 78.76 & 43.93 & 31.75 \\
HeteroFL ($r{=}1.0$, 10)   & 88.81 & 78.03 & 97.33 & 81.09 & 72.70 \\
HeteroFL ($r{=}1.0$, 50)   & 79.85 & 53.92 & 90.96 & 64.59 & 43.17 \\
HeteroFL ($r{=}1.0$, 100)  & 69.89 & 41.66 & 81.96 & 51.19 & 37.14 \\
\bottomrule
\end{tabular}}
\end{table*}
\begin{table*}[!t]
\centering
\caption{Impact of pruning ratio and strategy (server-side, client-side, adaptive) on COSMIC-FL (Sliced and Modular) on EuroSAT and RESISC45 with 50 satellites. Results report parameters (P, millions) and test accuracy (\%).}
\label{tab:ablation_pruning_full} \setlength{\tabcolsep}{4pt} 
\resizebox{\linewidth}{!}{
\begin{tabular}{c| c|cc|cc|cc|cc|cc|cc} 
\toprule \multirow{4}{*}{\rotatebox[origin=c]{90}{Model}} & \multirow{4}{*}{\rotatebox[origin=c]{90}{Pruning}} & \multicolumn{6}{c|}{EuroSAT} & \multicolumn{6}{c}{RESISC45} \\ \cmidrule(lr){3-8} \cmidrule(lr){9-14} & & \multicolumn{2}{c}{Server} & \multicolumn{2}{c}{Client} & \multicolumn{2}{c|}{Adaptive} & \multicolumn{2}{c}{Server} & \multicolumn{2}{c}{Client} & \multicolumn{2}{c}{Adaptive} \\ \cmidrule(lr){3-4} \cmidrule(lr){5-6} \cmidrule(lr){7-8} \cmidrule(lr){9-10} \cmidrule(lr){11-12} \cmidrule(lr){13-14} & & P & Acc & P & Acc & P & Acc & P & Acc & P & Acc & P & Acc \\ 
\midrule 
\multirow{6}{*}{\rotatebox[origin=c]{90}{\makecell{COSMIC-FL\\(Sliced)}}} & 0 & 12.4 & 87.56 & 12.4 & 87.56 & 12.4 & 87.56 & 12.5 & 64.07 & 12.5 & 64.07 & 12.5 & 64.07 \\ & 10 & 11.5 & 83.74 & 11.5 & 82.07 & 11.5 & 83.15 & 11.5 & 46.62 & 11.5 & 46.58 & 11.5 & 46.40 \\ & 30 & 8.2 & 81.15 & 8.2 & 81.04 & 8.4 & 81.93 & 8.3 & 44.91 & 8.3 & 44.31 & 8.5 & 46.96 \\ & 50 & 5.6 & 80.67 & 5.6 & 82.04 & 5.8 & 82.22 & 5.7 & 44.09 & 5.7 & 45.33 & 6.3 & 43.27 \\ & 70 & 3.5 & 80.59 & 3.5 & 79.67 & 4.0 & 80.89 & 3.6 & 41.20 & 3.6 & 40.36 & 3.9 & 42.18 \\ & 80 & 2.7 & 79.63 & 2.7 & 75.41 & 3.0 & 78.70 & 2.8 & 39.20 & 2.8 & 40.16 & 3.2 & 40.64 \\ 
\midrule 
\multirow{6}{*}{\rotatebox[origin=c]{90}{\makecell{COSMIC-FL\\(Modular)}}}& 0 & 19.8 & 87.37 & 19.8 & 87.37 & 19.8 & 87.37 & 19.9 & 66.44 & 19.9 & 66.44 & 19.9 & 66.44 \\ & 10 & 18.7 & 79.89 & 18.7 & 79.22 & 18.7 & 79.15 & 18.8 & 57.49 & 18.8 & 57.33 & 18.8 & 56.69 \\ & 30 & 16.4 & 80.11 & 16.4 & 79.78 & 16.6 & 80.41 & 16.5 & 56.64 & 16.5 & 56.40 & 16.5 & 57.96 \\ & 50 & 13.9 & 80.85 & 13.9 & 80.81 & 14.2 & 81.07 & 14.0 & 58.67 & 14.0 & 58.84 & 14.1 & 58.31 \\ & 70 & 11.2 & 81.00 & 11.2 & 81.63 & 11.3 & 81.33 & 11.3 & 59.82 & 11.3 & 59.98 & 11.4 & 59.73 \\ & 80 & 9.7 & 81.07 & 9.7 & 81.33 & 10.1 & 81.19 & 9.8 & 59.76 & 9.8 & 59.89 & 10.1 & 59.60 \\ 
\bottomrule \end{tabular}}
\end{table*}
\begin{table*}[!t]
\centering
\caption{Dense and adaptively pruned proposed models on the Jetson AGX Orin. Values are averaged over two independent runs.}
\label{tab:jetson_dense_pruned_supp}
\resizebox{\textwidth}{!}{
\begin{tabular}{lllrrrrr}
\toprule
Dataset & Impl. & Variant & Acc. (\%) & Ach. pruning (\%) & Latency (ms) & Throughput (img/s) & Peak memory (MB) \\
\midrule
EuroSAT & Modular & Dense & 86.22 & 0.00 & 10.59 & 94.43 & 83.45 \\
EuroSAT & Modular & Adaptive & 79.11 & 35.42 & 11.21 & 89.17 & 61.52 \\
EuroSAT & Sliced & Dense & 86.26 & 0.00 & 23.23 & 43.05 & 105.05 \\
EuroSAT & Sliced & Adaptive & 79.52 & 53.30 & 21.37 & 46.80 & 42.28 \\
\midrule
RESISC45 & Modular & Dense & 63.60 & 0.00 & 10.70 & 93.50 & 84.78 \\
RESISC45 & Modular & Adaptive & 58.87 & 39.81 & 11.23 & 89.07 & 55.01 \\
RESISC45 & Sliced & Dense & 60.96 & 0.00 & 23.23 & 43.05 & 105.26 \\
RESISC45 & Sliced & Adaptive & 43.67 & 51.52 & 21.52 & 46.47 & 44.48 \\
\bottomrule
\end{tabular}}
\end{table*}
\begin{table*}[!t]
\centering
\caption{Relative hardware effect of adaptive pruning with respect to the corresponding dense implementation.}
\label{tab:jetson_relative_pruning_supp}
\resizebox{\linewidth}{!}{
\begin{tabular}{llrrrr}
\toprule
Dataset & Impl. & Memory & Latency & Throughput & Power \\
\midrule
EuroSAT & Modular & $-26.28\%$ & $+5.90\%$ & $-5.57\%$ & $-3.96\%$ \\
EuroSAT & Sliced & $-59.75\%$ & $-7.98\%$ & $+8.70\%$ & $-10.75\%$ \\
RESISC45 & Modular & $-35.11\%$ & $+4.96\%$ & $-4.73\%$ & $-6.20\%$ \\
RESISC45 & Sliced & $-57.74\%$ & $-7.36\%$ & $+7.94\%$ & $-11.11\%$ \\
\bottomrule
\end{tabular}}
\end{table*}

\subsection{Total Uplink Communication across Training Phases}
Table~\ref{tab:total_comm} reports the total uplink communication across all training phases on EuroSAT under the reference schedule. Sparse-expert variants scale rotation communication by the measured expert-participation rates of Table~\ref{tab:sparse_expert_scales}. Since all experts are updated within each rotation round, communication scales with payload, rounds, and participating clients, rather than the number of experts. Two observations emerge. First, under full training, the staged pipeline may transmit more total data than FedAvg (3.6$\times$ for Modular and 1.4$\times$ for Sliced); its advantage lies in phase-specific communication, where satellites exchange only the parameters required by the active stage. Second, the communication reduction is achieved through sparse expert participation: at 100 clients, Modular-sparse and Sliced-sparse reduce total uplink by 47\% and 68\% compared with FedAvg, respectively, with larger gains as constellation size increases. Compared with sub-model baselines at $r{=}0.75$, Sliced-sparse becomes increasingly advantageous as participation becomes sparser.

\subsection{Sub-Model Baselines under Varying Capacity Fractions}
Table~\ref{tab:submodel_r_sweep} reports the full capacity-fraction sweep $r \in \{0.25, 0.5, 0.75, 1.0\}$ for FedDropout and HeteroFL under i.i.d.\ partitioning; the $r{=}0.75$ rows are repeated in the main paper. Accuracy increases monotonically with $r$ for both methods, and $r{=}1.0$ recovers full-model training, closely matching FedAvg. FedDropout degrades sharply at low $r$ (e.g., 5--33\% accuracy at $r{=}0.25$), since random sub-model sampling leaves each parameter undertrained, whereas HeteroFL degrades more gracefully because its deterministic width scaling trains a consistent sub-network every round. We therefore use $r{=}0.75$ as the strongest partial-capacity operating point when comparing against the proposed variants in the main paper.

\subsection{Structured Pruning on EuroSAT and RESISC45}

Table~\ref{tab:ablation_pruning_full} reports the full pruning study on both EuroSAT and RESISC45 (the main paper presents the EuroSAT columns). The RESISC45 results confirm the trends discussed in the main paper: differences between pruning strategies are more pronounced on RESISC45, the Modular variant remains competitive under strong pruning (70--80\%, up to 59.98\%), and the Sliced variant peaks at moderate ratios (46.96\% at 30\% adaptive pruning).

\section{Additional Jetson Hardware Results}\label{app:jetson_results}

This section provides additional embedded inference results on the NVIDIA Jetson AGX Orin. The measurements complement the server-side federated training evaluation by reporting latency, throughput, peak allocated CUDA memory, power, and temperature for dense and adaptively pruned checkpoints. Table~\ref{tab:jetson_dense_pruned_supp} reports the absolute measurements for the dense and adaptively pruned variants of the proposed modular and sliced implementations, while Table~\ref{tab:jetson_relative_pruning_supp} summarizes the relative hardware effect of adaptive pruning.

\subsection{\bf Embedded inference protocol}
Each final checkpoint is loaded and benchmarked independently on a single NVIDIA Jetson AGX Orin. The benchmark does not load all federated client models simultaneously and does not simulate a complete federated training round. Measurements use batch size one, 50 warm-up batches, and 500 timed batches. Two independent benchmark runs are performed for every checkpoint. Throughput is computed as the total number of processed images divided by the sum of the measured batch-inference times. Peak GPU memory is measured using \texttt{torch.cuda.max\_memory\_allocated()} after resetting CUDA peak-memory statistics. Board power and temperature are sampled through \texttt{tegrastats} during the measurement window.

{\color{teal}

}

{

}

Table~\ref{tab:jetson_relative_pruning_supp} summarizes the relative hardware effect of adaptive pruning with respect to the corresponding dense implementation. Adaptive pruning consistently reduces peak allocated CUDA memory and board power across both datasets and both expert implementations. The memory reduction is particularly large for the sliced implementation, reaching $59.75\%$ on EuroSAT and $57.74\%$ on RESISC45, while power decreases by about $11\%$. The effect on latency is implementation-dependent: pruning improves sliced-model latency by about $7$--$8\%$, but slightly increases modular-model latency by about $5$--$6\%$. This confirms that structured pruning provides clear memory and power benefits, whereas latency gains depend on the execution pattern and hardware behavior rather than parameter count alone.

We repeated the modular-versus-sliced baseline memory check three times. The sliced baseline consistently used more peak allocated CUDA memory than the modular baseline despite its smaller static parameter count. On EuroSAT, the modular baseline used 83.45\,MB and the sliced baseline used 105.05\,MB. On RESISC45, the corresponding values were 84.78\,MB and 105.26\,MB. This confirms that the runtime footprint is not determined by parameter count alone, but can also depend on activations, temporary tensors, path-specific indexing, and implementation-dependent CUDA workspaces.

A lightweight CUDA-event diagnostic further showed that the modular forward pass is largely accounted for by the explicit expert towers. In contrast, a substantial fraction of the sliced forward time was not attributed to the coarse instrumented \texttt{nn.Module} blocks. This observation suggests implementation-level overhead outside the main module stages, such as slicing, indexing, routing logic, tensor reorganization, or temporary allocation. The diagnostic should not be interpreted as isolating a single causal operation; a lower-level operator trace would be required to do so.


\end{document}